\documentclass[final,5p,times,twocolumn]{elsarticle}

\usepackage{graphicx}
\usepackage{subfig}
\usepackage{multirow}
\usepackage{slashbox}
\usepackage{algorithm}
\usepackage{epsfig}

\usepackage{amssymb}

\usepackage[nodots]{numcompress}
\usepackage{epstopdf}

\DeclareGraphicsExtensions{.eps,.mps,.pdf,.jpg,.png}

\usepackage{lineno}

\journal{Computers \& Graphics}

\begin{document}

\begin{frontmatter}



\title{CNN Feature boosted SeqSLAM for Real-Time Loop Closure Detection}

 \author[label1,label2]{Dongdong Bai}
 \author[label1,label2]{Chaoqun Wang}
 \author[label1,label2]{Bo Zhang\corref{cor1}}
 \author[label1,label2]{Xiaodong Yi}
 \author[label1,label2]{Xuejun Yang}

\address[label1]{College of Computer, National University of Defense Technology, ChangSha, 410073, China}
\address[label2]{State Key Laboratory of High Performance Computing, National University of Defense Technology, ChangSha, 410073, China}
\cortext[cor1]{Corresponding author: zhangbo10@nudt.edu.cn}

\begin{abstract}
Loop closure detection (LCD) is an indispensable part of simultaneous localization and mapping systems (SLAM); it enables robots to produce a consistent map by recognizing previously visited places. When robots operate over extended periods, robustness to viewpoint and condition changes as well as satisfactory real-time performance become essential requirements for a practical LCD system.

This paper presents an approach to directly utilize the outputs at the intermediate layer of a pre-trained convolutional neural network (CNN) as image descriptors. The matching location is determined by matching the image sequences through a method called \emph{SeqCNNSLAM}. The utility of SeqCNNSLAM is comprehensively evaluated in terms of viewpoint and condition invariance.  Experiments show that SeqCNNSLAM outperforms state-of-the-art LCD systems, such as SeqSLAM and Change Removal, in most cases. To allow for the real-time performance of SeqCNNSLAM, an acceleration method, \emph{A-SeqCNNSLAM}, is established. This method exploits the location relationship between the matching images of adjacent images to reduce the matching range of the current image. Results demonstrate that acceleration of 4-6 is achieved with minimal accuracy degradation, and the method's runtime satisfies the real-time demand. To extend the applicability of A-SeqCNNSLAM to new environments, a method called \emph{O-SeqCNNSLAM} is established for the online adjustment of the parameters of A-SeqCNNSLAM.
\end{abstract}


\end{frontmatter}


\section{Introduction}
Large-scale navigation in a changing environment poses a significant challenge to robotics, because during this process, a robot inevitably encounters severe environmental changes. Such changes are mainly divided into condition and viewpoint. Condition change is caused by changes in the external environment, such as changes in illumination, weather, and even season. The appearance of a single area could differ depending on the external environment. Meanwhile, robots encountering a viewpoint change may view the same area from various perspectives as they move around.

Aside from robustness to viewpoint and condition change, real-time performance is another inevitable re-quirement for loop closure detection (LCD). Evidently, a robot should be able to determine without a high overhead cost whether its current location has been visited. However, a robot is equipped with a computer, the computing power of which is close to that of a personal computer. The computer usually implements other robotic applications simultaneously, along with the LCD algorithm. Therefore, the LCD algorithm should not have significant computing requirements.

Many researchers have successfully addressed view-point and condition changes. For example, bag of words\cite{Sivic-2003-p1470}  was introduced into FAB-MAP\cite{Cummins-2008-p647}. Excellent performance was achieved with regard to viewpoint change, and the method has become one of the state-of-the-art approaches for LCD based on single-image matching. Recently, Milford et al. proposed a method of matching sequence images called SeqSLAM\cite{Milford-2012-p1643} and achieved improved robustness against condition change.

The features used by these approaches are hand crafted and designed by experts with domain-specific knowledge. However, robots may face various complexes and uncertain environments during localization. Hence, considering all the factors that affect the performance of LCD is difficult. Such as, SeqSLAM is only condition invariance, but it shows poor robustness against viewpoint change.

Recently, convolutional neural networks (CNN) has been achieved great success\cite{Krizhevsky-2012-p1097}\cite{Chatfield-2014-p} and received much interest in applying CNN features to robotic fields\cite{Hou-2015-p2238}\cite{Suenderhauf-2015-p4297}\cite{Suenderhauf-2015-p}\cite{Lowry-2016-p1}. Hou et al.\cite{Hou-2015-p2238} and  Sunderhauf et al.\cite{Suenderhauf-2015-p4297} were pioneers of these researchers. Instead of using hand-crafted features,  they respectively analyzed the utility of each layer of two pre-trained CNN models with an identical architecture---places-CNN\cite{Zhou-2014-p487} and AlexNet\cite{Krizhevsky-2012-p1097} (the architecture is shown in Fig.\ref{figure1}). Their results showed that conv3 and pool5 are representative layers that demonstrate beneficial condition and viewpoint invariance.

However, CNN descriptors, such as conv3 and pool5, exhibit strong robustness to either viewpoint change or condition change. Simultaneously demonstrating robustness to both condition and viewpoint change remains an open problem. Furthermore, the dimensions of CNN descriptors are high, and the similarity degree between images is measured by their features' Euclidean distance. So the operation of a CNN-based LCD algorithm requires a large amount of computation, which poses a difficulty in meeting the real-time demand for robots.

In this paper, we present a robust LCD method by fusing the CNN features and sequences matching method. And, we optimize its real-time performance by reducing image matching range. Our contributions are two fold:
\begin{itemize}
	\item First, we present \emph{SeqCNNSLAM} to combine CNN descriptors and sequence matching to enable the algorithm to cope with viewpoint and condition change simultaneously. Comprehensive experiments are conducted to demonstrate that SeqCNNSLAM exhibits more robust performance than several state-of-the-art methods, such as SeqSLAM\cite{Lowry-2015-p} and Change Removal\cite{Milford-2012-p1643}.
	\item Second, we reduce the computational complexity of SeqCNNSLAM to determine matching images for the current image by limiting the matching range based on the sequential information underlined between images. The approach is called \emph{A-SeqCNNSLAM}, which obtains 4-6 times of acceleration and still presents a performance that is comparable to that of SeqCNNSLAM. Meanwhile, we provide an approach for the online adjustment of several parameters in A-SeqCNNSLAM. The approach is called \emph{O-SeqCNNSLAM}; it enhances the practical performance of the algorithm.
\end{itemize}

The paper proceeds as follows. Section II provides a brief background of LCD as well as CNN development and status quo. The image descriptors, LCD performance metrics, and datasets used in the experiments are provided in Section III. Sections IV presents the implementation of algorithms and the tests conducted on SeqCNNSLAM. In Section V, we provide the detailed introduction on A-SeqCNNSLAM and O-SeqCNNSLAM, and do comprehensive research about their performance and runtime. Section VI presents the results, conclusions, and suggestions for future work.

\section{Related Work}
This section provides a brief introduction of LCD, CNN, and several of the latest studies on applying pre-trained CNN descriptors to LCD.
\subsection{Loop Closure Detection}
The capability to localize and generate consistent maps of dynamic environments is vital to long-term robot autonomy\cite{Barfoot-2013-p1609}. LCD is a technique to determine the locations of a loop closure of a mobile robot's trajectory, which is critical to building a consistent map of an environment by correcting the localization errors that accumulate over time. Given the change in the environment and movement of the robot, images corresponding to the same place collected by a robot may present an entirely different appearance because of the existence of condition and viewpoint change.

To cope with these challenges, researchers have designed many methods for various situations. Speeded-up robust features (SURF)\cite{Bay-2006-p404} are typical examples. Since the bag-of-words \cite{Sivic-2003-p1470} model was proposed, SURF\cite{Bay-2006-p404} descriptors have been widely used in LCD. The bag-of-words\cite{Sivic-2003-p1470} model was first introduced into LCD by FAB-MAP\cite{Cummins-2008-p647}, which is based on the SURF\cite{Bay-2006-p404} descriptor. Considering the invariance properties of SURF\cite{Bay-2006-p404} in generating bag-of-words\cite{Sivic-2003-p1470} descriptors and that the bag-of-words model ignores the geometric structure of the image it describes, FAB-MAP\cite{Cummins-2008-p647} demonstrates robust performance in viewpoint change and has become a state-of-the-art algorithm based on a single image match in LCD research.

Instead of searching the single location most similar to the current image, Milford et al. proposed an approach that calculates the best candidate matching the location based on a sequence of images. Their approach, which was coined SeqSLAM\cite{Milford-2012-p1643}, achieves remarkable results with regard to coping with condition change and even season change\cite{Niko-2013-p}. Searching for matching sequences is deemed the core of SeqSLAM\cite{Niko-2013-p}.
\subsection{Convolutional Neural Networks}
Since Alex et al.\cite{Krizhevsky-2012-p1097} became the champion of the Imagenet Large Scale Visual Recognition Competition 2012 (ILSVRC2012), algorithms based on CNN have dominated over traditional solutions that use hand-crafted features or operate on raw pixel levels in the computer vision and machine learning community\cite{babenko2014neural}\cite{Wan-2014-p157}. Much interest has been devoted to applying convolutional network features to robotic fields, such as visual navigation and SLAM\cite{Hou-2015-p2238}\cite{Suenderhauf-2015-p4297}\cite{Suenderhauf-2015-p}\cite{Lowry-2016-p1}. The methods of using CNN for robotic applications fall into two categories: training a CNN for a robotic application or directly using outputs of several layers in a pre-trained CNN for related robotic applications.

Directly training a CNN for a specific application is a complex process not only because of the need for a large amount of data in this field, but also because obtaining a CNN with a remarkable performance demands that researchers have plenty of experience to adjust the architecture and parameters of the CNN. That is, many tricks can be applied to train a CNN, and this presents great difficulties in using the CNN for robotics.

Given the versatility and transferable property of CNN\cite{Razavian-2014-p512}, although they were trained for a highly specific target task, they can be successfully used to solve different but related problems.

Hou et al.\cite{Hou-2015-p2238} investigated the feasibility of the public pre-trained model (Places-CNN)\cite{Zhou-2014-p487} as an image descriptor generator for visual LCD. Places-CNN was implemented by the open-source software Caffe\cite{Jia-2014-p675} and trained on the scene-centric dataset Places\cite{Zhou-2014-p487}, which contains over 2.5 million images of 205 scene categories for place recognition. The researchers comprehensively analyzed the performance and characteristics of descriptors generated from each layer of the Places-CNN model and demonstrated that the conv3 layer presents some condition invariance, whereas the pool5 layer presets some viewpoint invariance.

Compared with hand-crafted features, CNN descriptors can easily deal with complex and changeable environments and do not require researchers to have much domain-specific knowledge.

\begin{figure*}
	\includegraphics[width=1\linewidth]{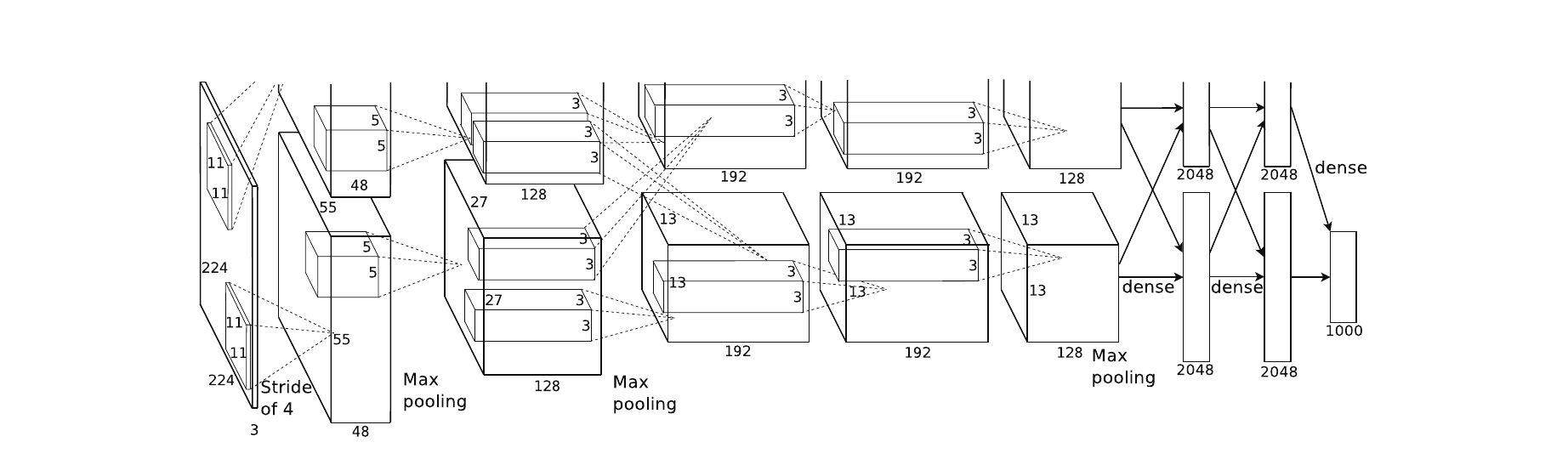}\caption{Architecture of the Places-CNN/AlexNet model\$\textasciicircum{}\{{[}6{]}\}\$.}
	\label{figure1}
	
\end{figure*}

\begin{table*}
	\caption{Dimensions of each layer of the Places-CNN/AlexNet model.}

	\begin{tabular}{c||c|c|c|c|c|c|c|c||c|c|c}
		\hline
		& \multicolumn{8}{c||}{Convolutional} & \multicolumn{3}{c}{Fully-Connected}\tabularnewline
		\hline
		Layer & CONV1 & POOL1 & CONV2 & POOL2 & CONV3 & CONV4 & CONV5 & POOL5 & FC6 & FC7 & FC8\tabularnewline
		\hline
		Dimension & 290400 & 69984 & 186624 & 43264 & 64896 & 64896 & 43264 & 9216 & 4096 & 4096 & 1000\tabularnewline
		\hline
	\end{tabular}
	\label{table1}
\end{table*}

\subsection{Combination of SeqSLAM with CNN Descriptors}
Given the advantages of SeqSLAM and CNN, Lowry et al. considered combining the sequence matching method with CNN descriptors to fuse their respective advantages and construct an LCD system that is robust to both viewpoint and condition changes. Their method is called Change Removal\cite{Lowry-2015-p}. Change Removal involves two main processes. First, it removes a certain number of the earliest principal components of images to remove information on images that is related to condition change. The rest of the principal components of images are used as input for CNN to obtain a robust feature against viewpoint and condition changes. However, the performance of Change Removal\cite{Sivic-2003-p1470} depends largely on a dataset-specific parameter, that is, the number of principal components of images to be removed. Therefore, selecting the setting for unseen scenes and environments is difficult.

In this study, we present a new means to maximize both CNN descriptors and the sequence matching method. Preprocessing of images is not needed, and the images are directly used as input to CNN. Compared with Change Removal\cite{Lowry-2015-p}, the proposed method is more general and does not depend on any dataset-specific parameters.

\section{Preliminaries}
\subsection{Obtain CNN descriptors}
For a pre-trained CNN model, the output vector of each layer can be regarded as an image descriptor $\tilde{X}$. A descriptor of each layer can be obtained by traveling through the CNN networks. Before using  $\tilde{X}$, it is normalized to become unit vector $X$ according to the following equation:
\begin{equation}
\left(\frac{\tilde{x}_{1}}{\sqrt{\sum_{i=1}^{d}\tilde{x}_{i}^{2}}},\cdots,\frac{\tilde{x}_{d}}{\sqrt{\sum_{i=1}^{d}\tilde{x}_{i}^{2}}}\right)\rightarrow\left(x_{1,\cdots,}x_{d}\right),
\end{equation}
where  $(\tilde{x}_{1,\cdots,}\tilde{x}_{d})$ is a descriptor $\tilde{X}$ with $d$ dimensions and $(x_{1,\cdots,}x_{d})$ is normalized descriptor $X$ with an identical dimension as $\tilde{X}$.

Algorithm \ref{algorithm1} shows the process of obtaining CNN descriptors.
\begin{algorithm}
	\caption{Obtain CNN descriptors}

	\textbf{Require:} $\ensuremath{S=\left\{ \left(x_{i},y_{i}\right)\right\} _{i=1}^{N}}$:dataset
	for LCD containing $N$ images, $x_{i}$: input images,$y_{i}$: the
	ground truth of matching image's serial number for$x_{i}$; $Normalize(Y)$:
	normalize vector $Y$ based on Eq.(1).
	
	\textbf{Ensure:} $X^{1}=\left\{ X_{i}^{1}\right\} _{i=1}^{N}$: conv3
	descriptors of $S$, $X^{2}=\left\{ _{i}^{2}\right\} _{i=1}^{N}$:
	pool5 descriptors of $S$.
	
	1$\quad$for $i=1:N$
	
	2$\quad$$\quad$$\quad$Put $x_{i}$ into Places-CNN to get the output
	of conv3 $\left(\tilde{X}_{i}^{1}\right)$and pool5 $\left(\tilde{X}_{i}^{2}\right)$;
	
	3$\quad$$\quad$$\quad$$X_{i}^{1}=Normalize\left(\tilde{X}_{i}^{1}\right)$;
	
	4$\quad$$\quad$$\quad$$X_{i}^{2}=Normalize\left(\tilde{X}_{i}^{2}\right)$;
	
	5$\quad$end for
	\label{algorithm1}
\end{algorithm}

\subsection{Performance Measures}
The performance of the LCD algorithm is typically evaluated according to precision, recall metrics and precision-recall curve. The matches consistent with the ground truth are regarded as true positives (TP), the matches inconsistent with the ground truth are false positives (FP), and the matches erroneously discarded by the LCD algorithm are regarded as false negative matches (FN). Precision is the proportion of matches recognized by the LCD algorithm as TP matches, and recall is the proportion of TP matches to the total number of actual matches in the ground truth, that is,

\begin{equation}
Precision=\frac{TP}{TP+FP},
\end{equation}

\begin{equation}
Recall=\frac{TP}{TP+FN}.
\end{equation}

For LCD,  the maximum recall at 100\% precision is also an important  performance indicator; it is widely used by many researchers to evaluate LCD algorithms and is also used in our subsequent experiments. Although this criterion may cause the algorithms to reject several correct matches in the ground truth,  the cost of adopting an FP in the LCD system is extremely large and often results in an erroneous map in the SLAM system. For the subsequent test, we accepted a proposed match if it was within two frames of the ground truth.
\subsection{Datasets used in the Evaluation}
In the subsequent experiments, we used two datasets with different properties to evaluate the utility of the algorithms.
\subsubsection{Nordland Dataset(winter-spring)}
\begin{figure}
	
	\subfloat{\includegraphics[width=0.25\columnwidth]{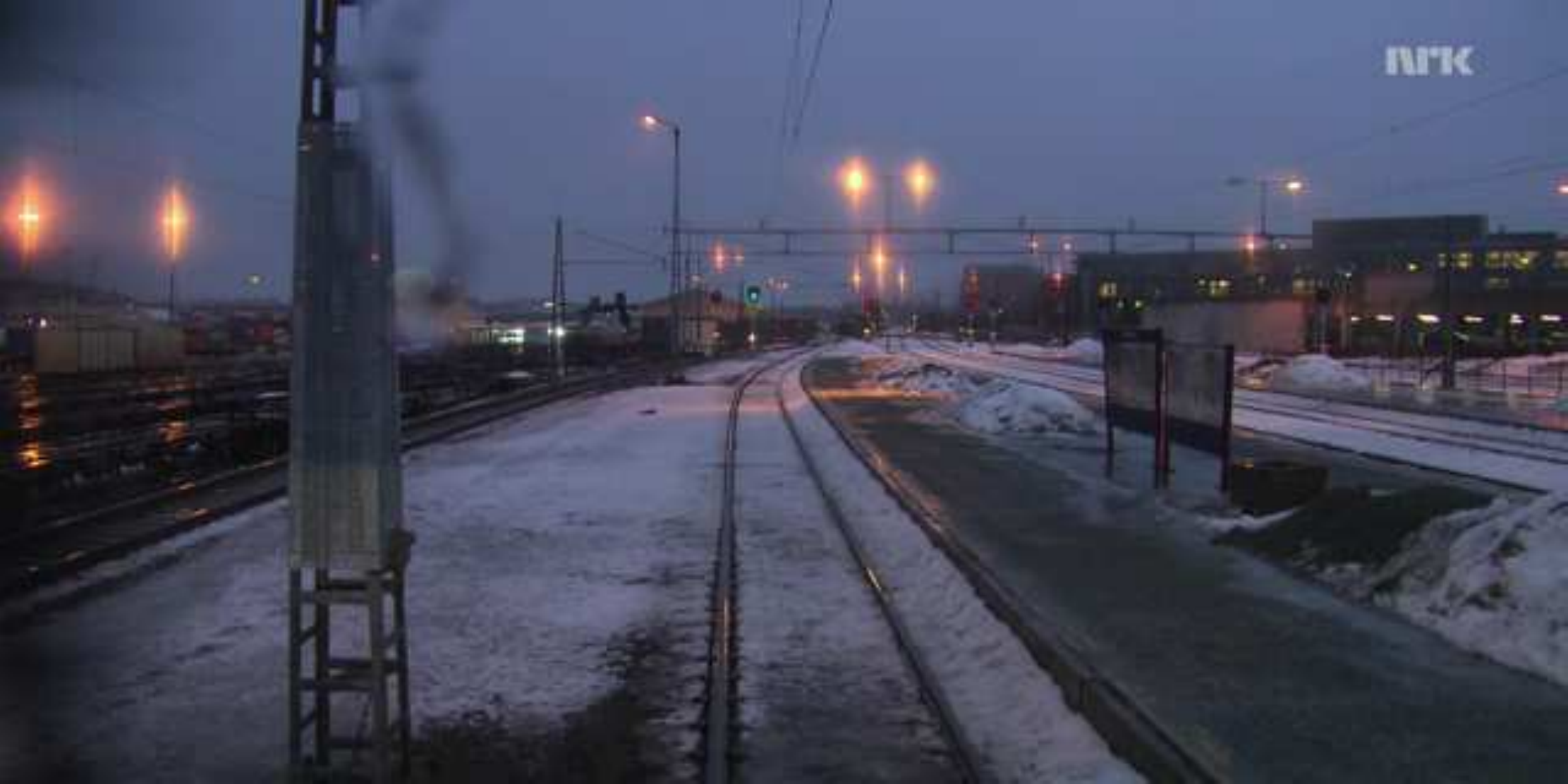}}\subfloat{\includegraphics[width=0.25\columnwidth]{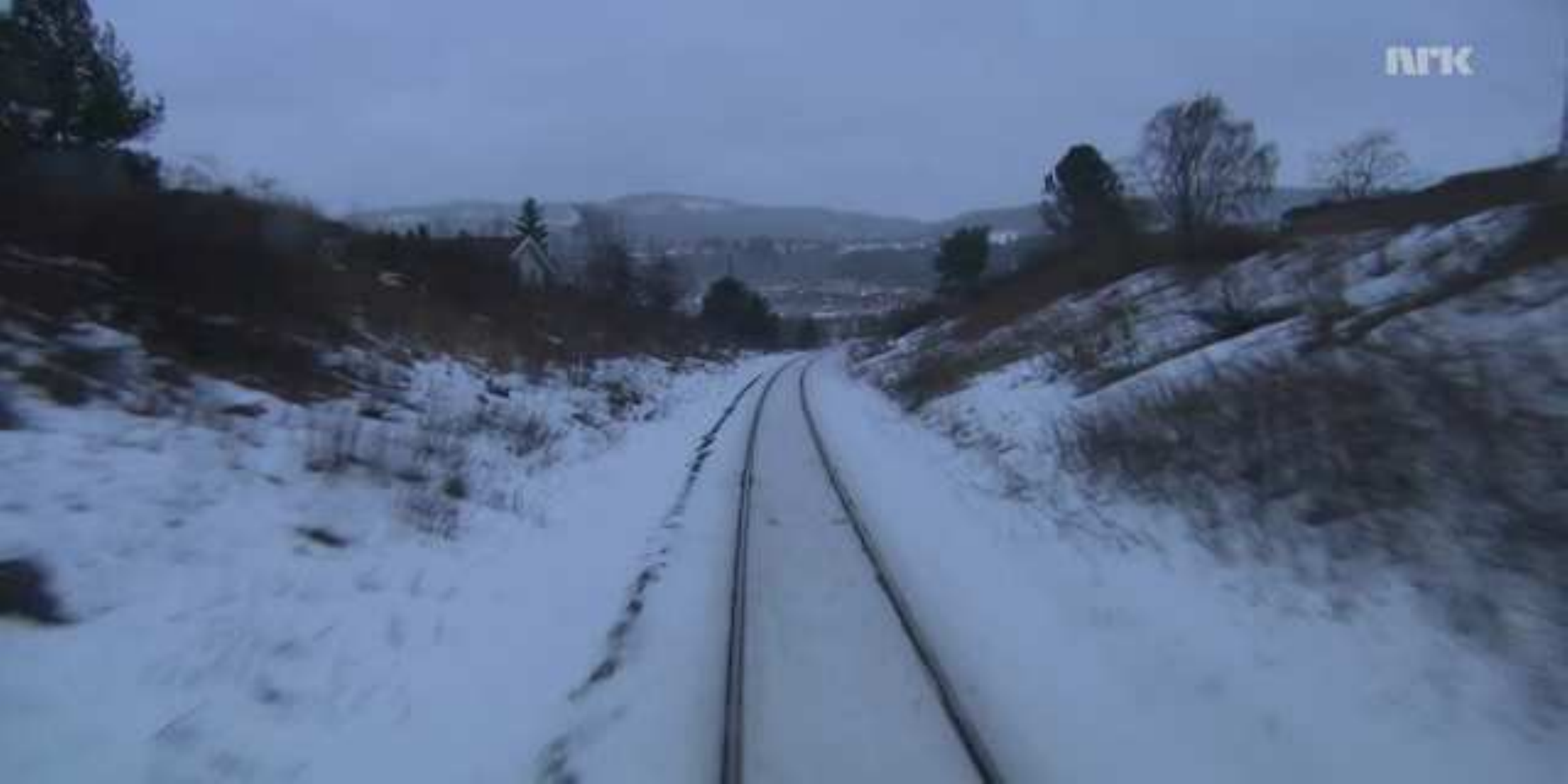}}\subfloat{\includegraphics[width=0.25\columnwidth]{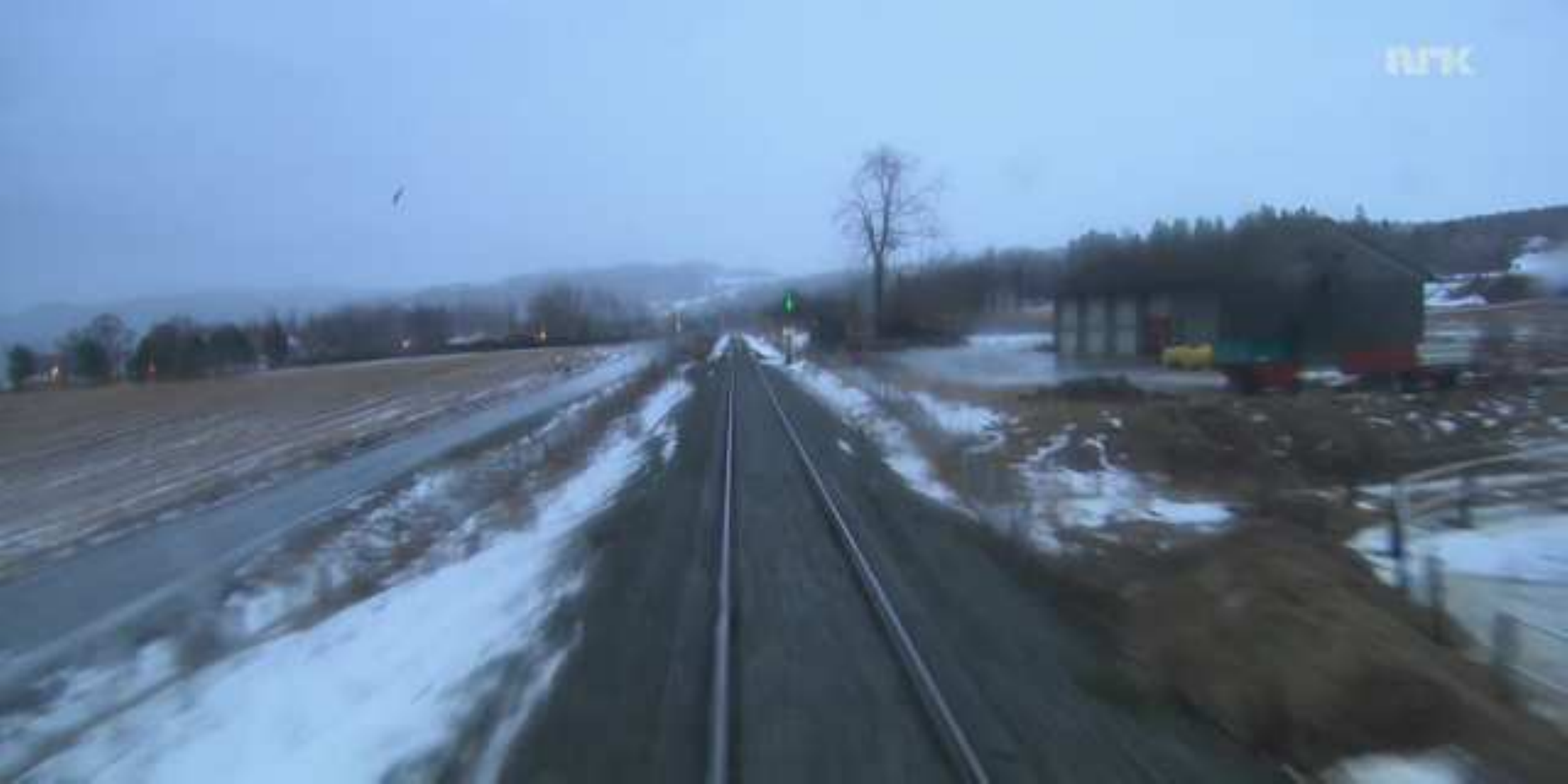}}\subfloat{\includegraphics[width=0.25\columnwidth]{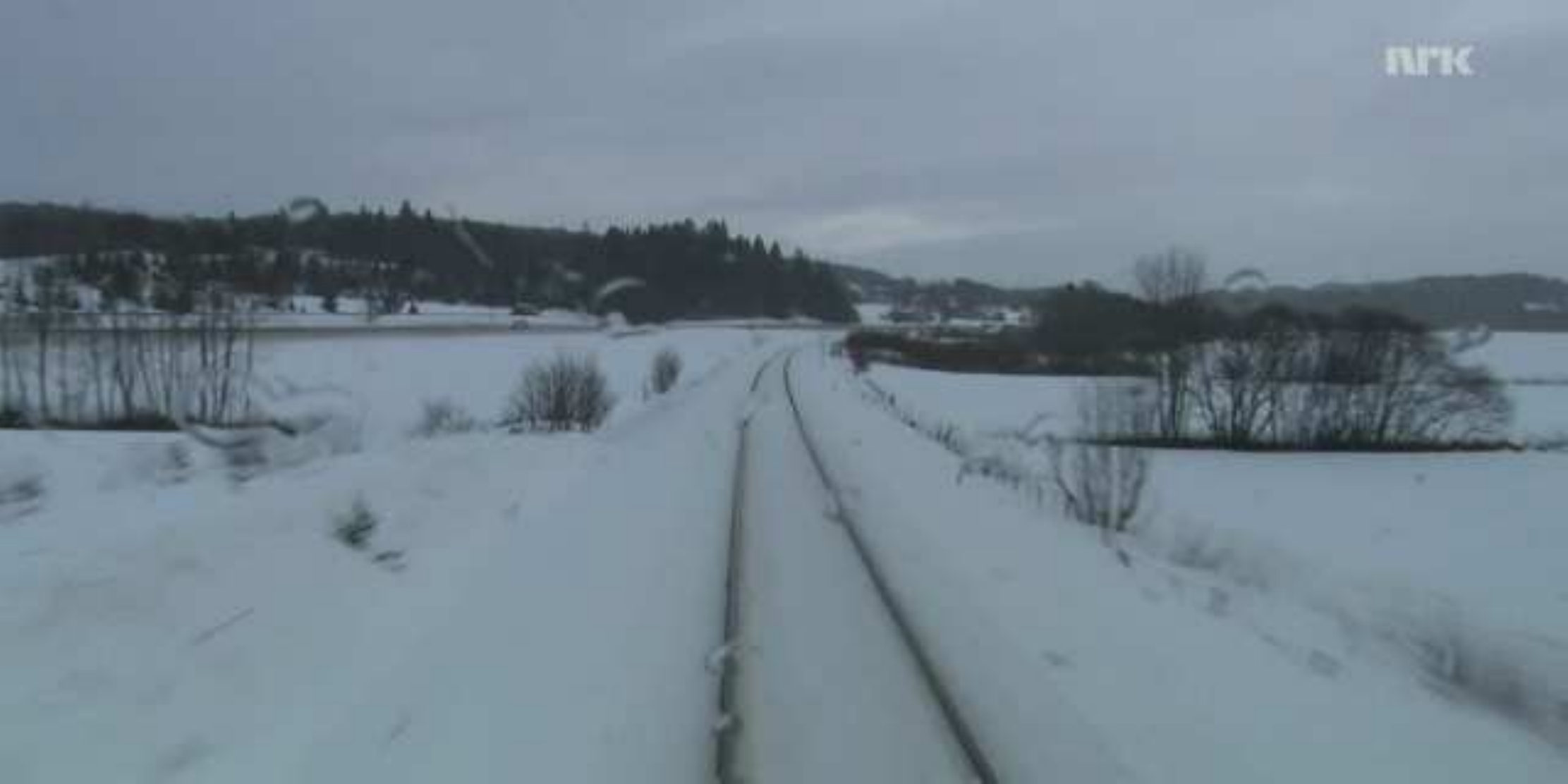}}
	
	\subfloat{\includegraphics[width=0.25\columnwidth]{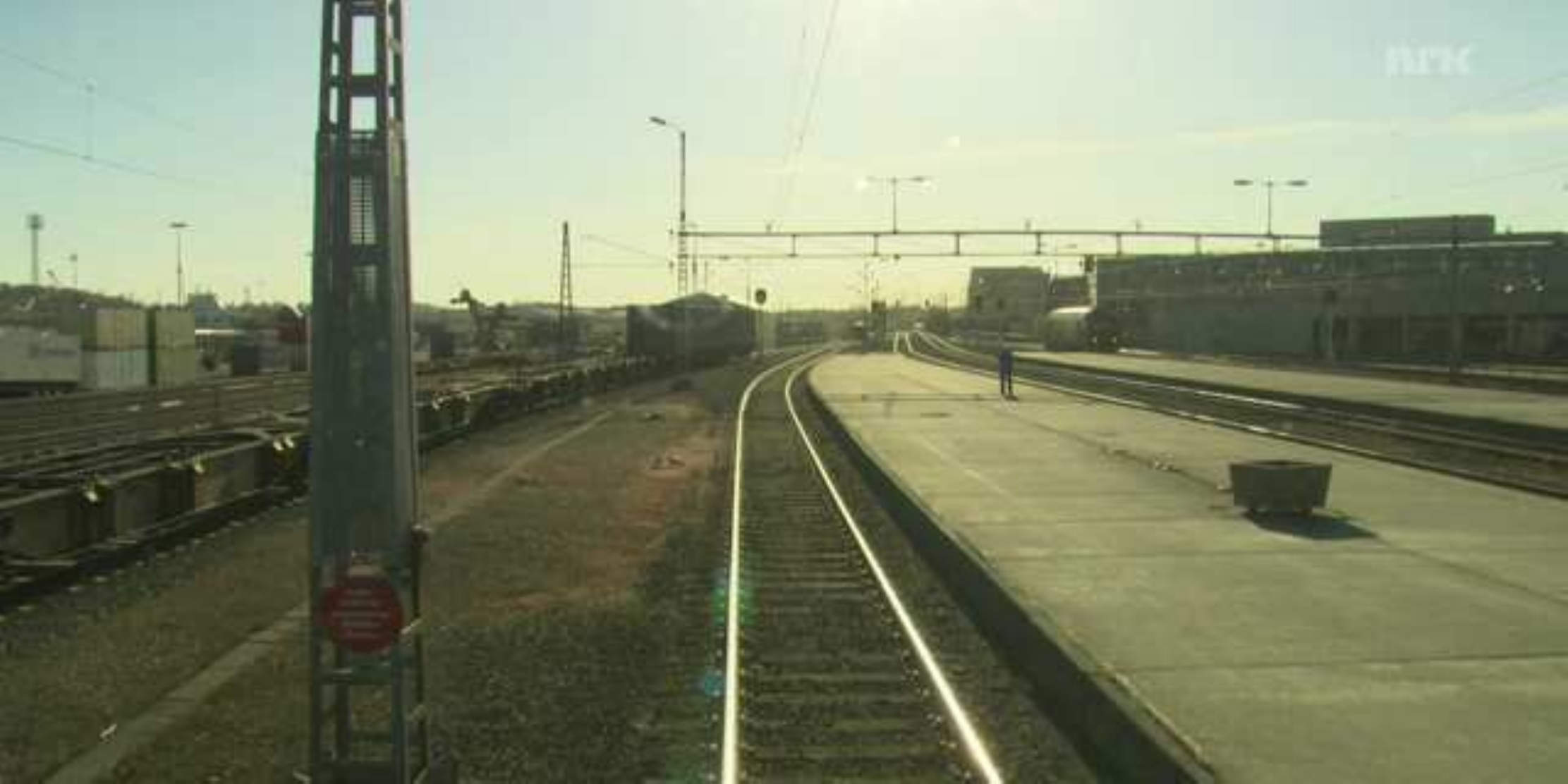}}\subfloat{\includegraphics[width=0.25\columnwidth]{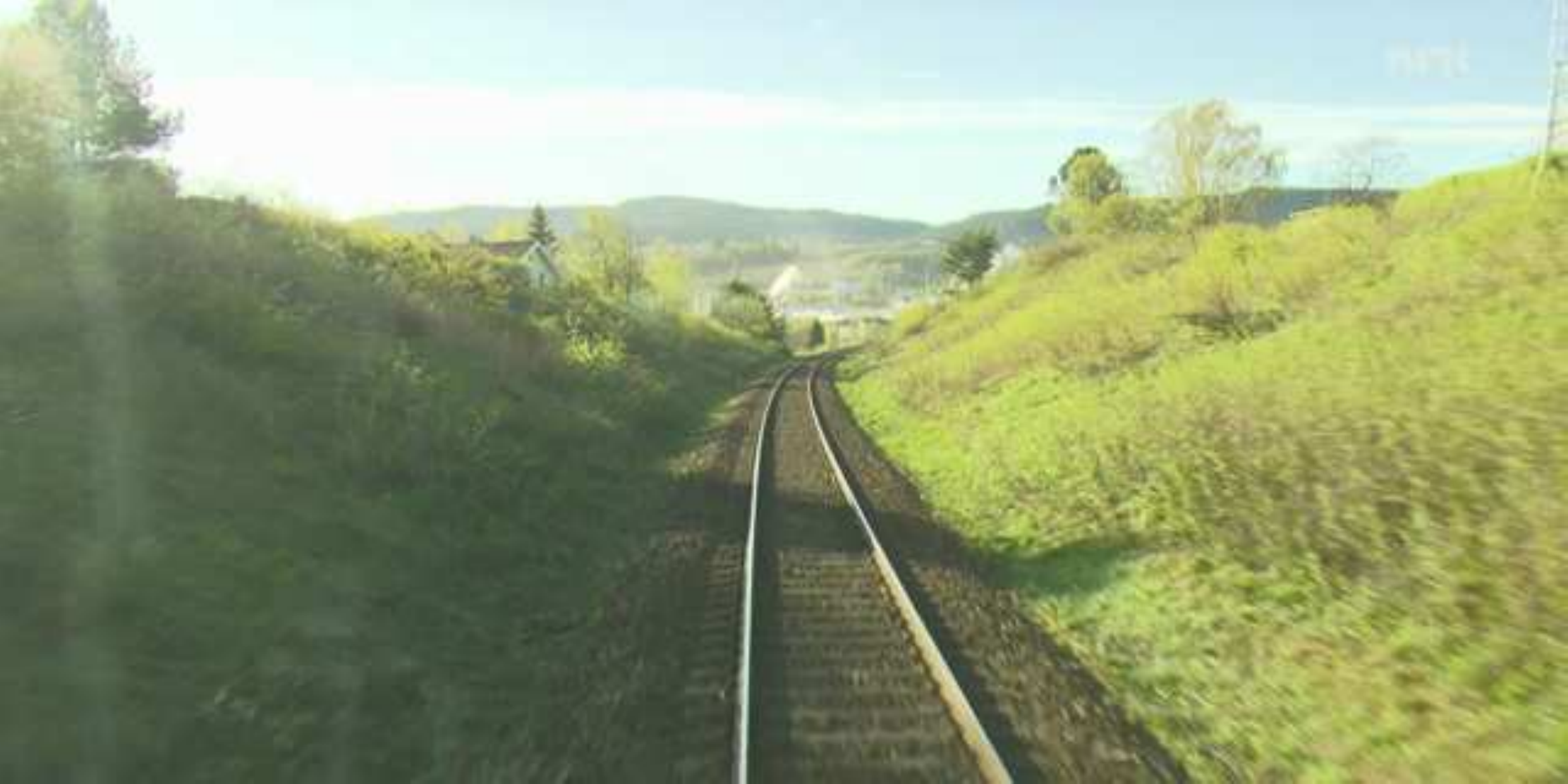}}\subfloat{\includegraphics[width=0.25\columnwidth]{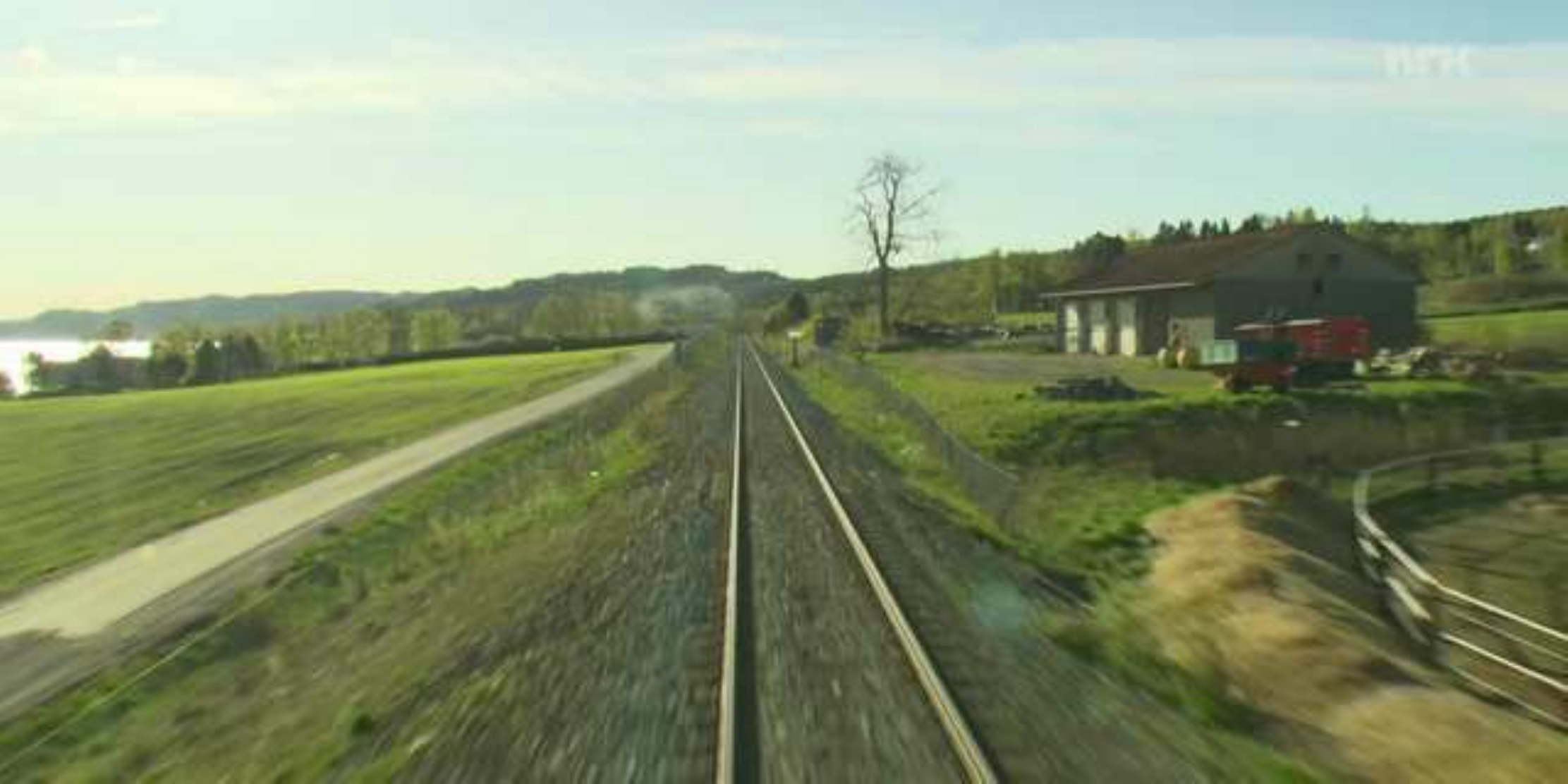}}\subfloat{\includegraphics[width=0.25\columnwidth]{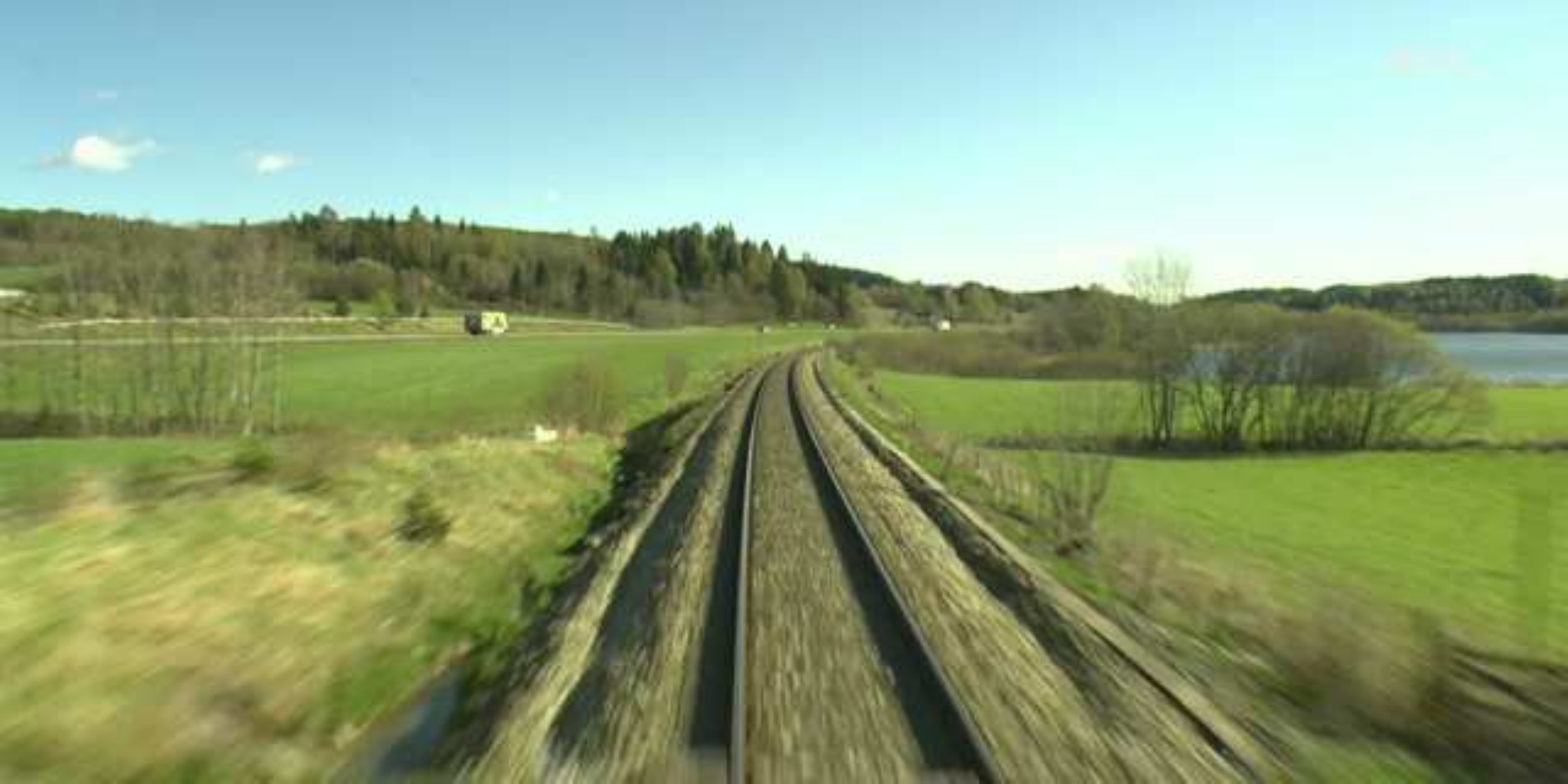}}
	\caption{Sample Nordland images from matching locations: winter (top) and spring
	(bottom).}
\label{figure2}
\end{figure}

The Nordland dataset was produced by extracting still frames from a TV documentary ``Nordlandsbanen-Minutt for Minutt" by the Norwegian Broadcasting Corporation. The TV documentary records the 728 km train ride in northern Norway from the same perspective in the front of a train in four
different seasons for 10 hours. The dataset has been used to evaluate the performance of OpenSeqSLAM\cite{Niko-2013-p}, an open source implementation of SeqSLAM that copes with season changes.
The dataset captured in four different seasons exhibits simple condition changes because of the same running path of the train and orientation of the camera. Fig.\ref{figure2} shows an intuitionistic impression of the severe appearance change between seasons. As illustrated in the figure, a severe seasonal change from full snow cover in winter to over-green vegetation in spring occurs and is the most severe change among all pairs, such as spring to summer. Hence, we adopted the spring and winter seasons of the Nordland dataset for our subsequent experiments.

The TV documentary is recorded at 25 fps with a resolution of 1920*1080, and GPS information is recorded in conjunction with the video at 1 Hz. The videos and GPS information are publicly available online$^1$  \footnotetext{\footnotesize$^1$https://nrkbeta.no/2013/01/15/nordlandsbanen-minute-by-minute-season-by-season/}. The full HD recordings have been time synchronized; the position of the train in an arbitrary frame from one video corresponds to a frame with the same serial number in any of the other three videos.

For our experiments, we extract image frames at a rate of 1 fps from the video start to time stamps 1:30 h. We down-sample frames to 640*320 and excluded all images obtained inside tunnels or when the train was stopped. By now, we obtained 3476 images for each season.
\subsubsection{Gardens Point Dataset}
\begin{figure}
	\subfloat{\includegraphics[width=0.33\columnwidth]{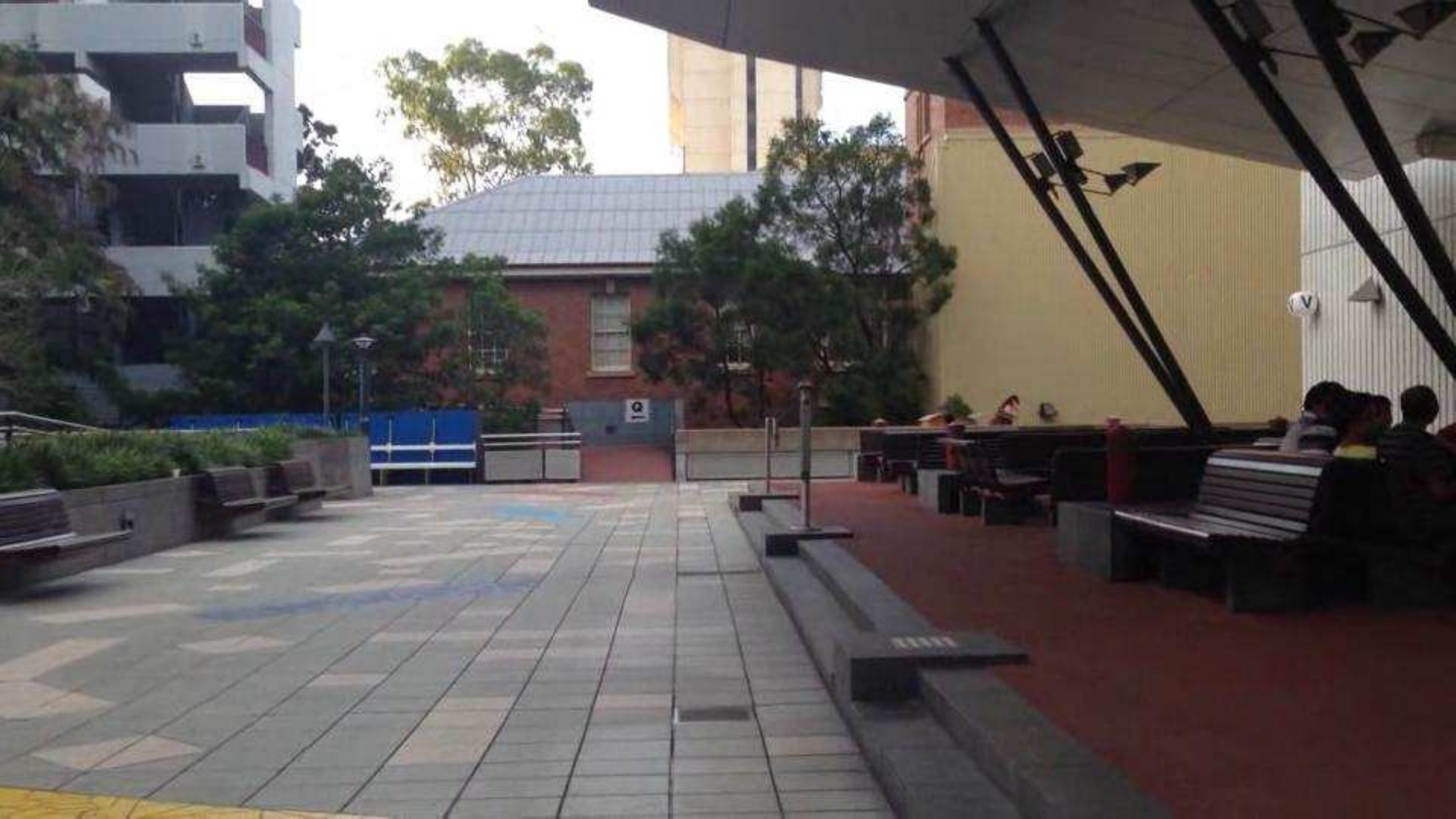}}\subfloat{\includegraphics[width=0.33\columnwidth]{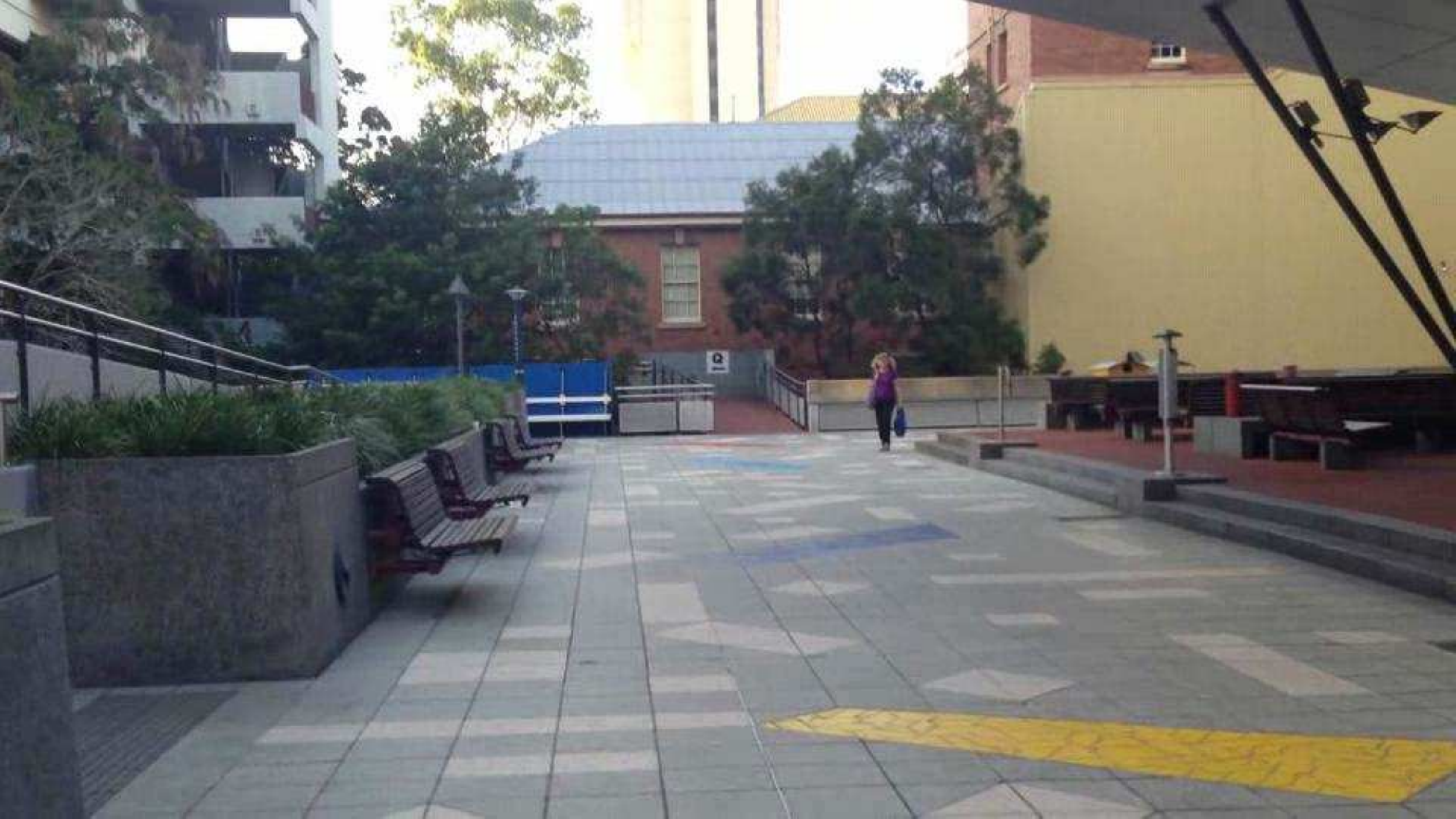}}\subfloat{\includegraphics[width=0.33\columnwidth]{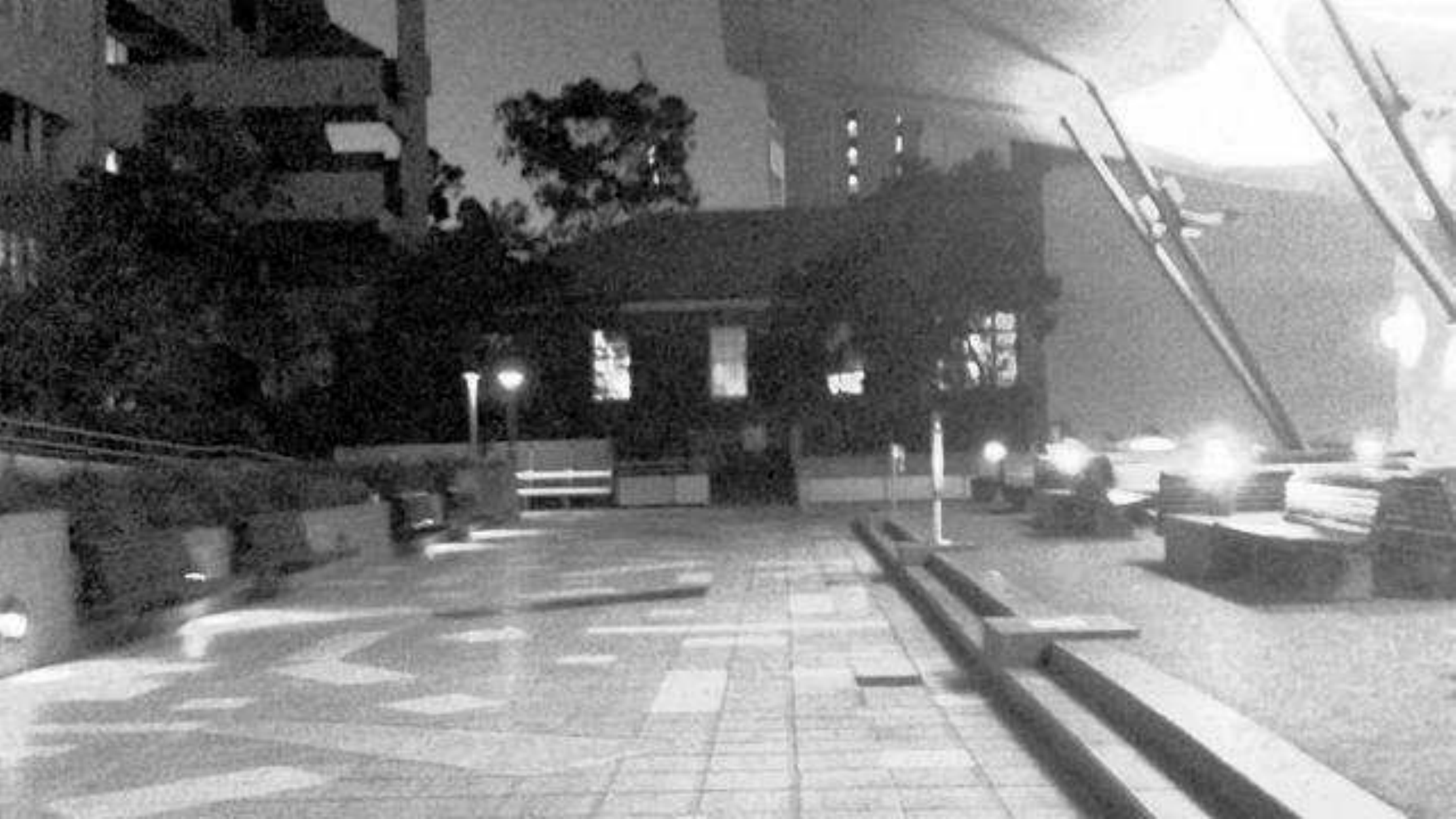}}
	
	\caption{Sample Gardens Point images from matching locations: day-right (left),
		day-left (middle) and night-right (right).}
	\label{figure3}
	
\end{figure}

The Gardens Point dataset was recorded on a route through the Gardens Point Campus of Queensland University of Technology. The route was traversed three times: twice during the day and once at night. One day route (day-left) was traversed on the left-hand side of the path, and the other day route (day-right) and the night route (night-right) were traversed on the right-hand aspect of the road.  Two hundred images were collected from each traversal, and an image name corresponds to the location in each traversal. Thus, the dataset exhibits both condition and viewpoint changes, as illustrated in Fig.\ref{figure3}. The dataset is publicly available online$^2$ \footnotetext{\footnotesize$^2$https://wiki.qut.edu.au/display/cyphy/Day+and+Night+with+Lateral
+Pose+Change+Datasets}.

\section{SeqCNNSLAM Loop Closure Detection}
This section presents the comprehensive design of SeqCNNSLAM, which realizes a robust LCD algorithm combining  CNN descriptors and SeqSLAM\cite{Milford-2012-p1643}.

SeqSLAM\cite{Milford-2012-p1643} has been described by Milford et al. in detail. For comparison with the proposed approach, the important ideas and algorithmic steps of SeqSLAM\cite{Milford-2012-p1643} are summarized below. SeqSLAM\cite{Milford-2012-p1643} mainly involves \emph{three steps}. First, in the preprocessing step, incoming images are drastically down-sampled to, for example, 64*32 pixels, and divided into patches of 8*8 pixels. The pixel values are then normalized between 0 and 255. The image difference matrix is obtained by calculating all the preprocessed images using the sum of absolute differences. Second, the distance matrix is locally contrast enhanced. Finally, when looking for a match to a query image, SeqSLAM\cite{Milford-2012-p1643} performs a sweep through the contrast-enhanced difference matrix to find the best matching sequence of frames based on the sum of \emph{sequence differences}. The process to calculate the sequence differences is illustrated in Algorithm \ref{algorithm2}.

\begin{algorithm}
	\caption{Cal-Seq-Dif($N$,$q$,$ds$)}

	\textbf{Require: }$ds$: sequence length; $n$: current image's serial
	number; $D\in R^{N\times N}$: difference matrix, $D_{ij}$ : an element
	of $D$; $V$: trajectory speed; $q$: middle image's serial number
	of a matching sequence.
	
	\textbf{Ensure: $sum_{nq}$}: the sum of $q$-th sequence differences
	for $n$-th image.
	
	1$\quad$for $i=0:ds$
	
	2$\quad$$\quad$$\quad$$sum_{nq}=sum_{nq}+D_{V*\left(q-\frac{ds}{2}+i\right),n-\frac{ds}{2}+i}$
	
	3$\quad$end for
	
	4$\quad$return $sum_{nq}$
	\label{algorithm2}
\end{algorithm}

SeqSLAM\cite{Milford-2012-p1643} has a large number of  parameters that control the behavior of the algorithm and the quality of its results. The parameter \emph{ds} is presumably the most influential one; it controls the length of image sequences that are considered for matching. SeqSLAM\cite{Milford-2012-p1643} is expected to perform better with the increase in $ds$ because longer sequences are more distinct and less likely to result in FP matches.

The remarkable performance of SeqSLAM\cite{Milford-2012-p1643} under condition change demands a relatively stable viewpoint in the different traversals of the environment, which are caused by directly using the sum of absolute difference to measure the difference between images. But the matching sequence, which is the core of SeqSLAM, is a significant contribution of SeqSLAM to LCD\cite{Niko-2013-p}.

Several achieved CNN layers outperform hand-crafted descriptors in terms of condition and viewpoint changes, so we developed the method \emph{SeqCNNSLAM} to combine CNN descriptors that are invariant to viewpoint changes with the core of SeqSLAM\cite{Niko-2013-p}.

In our work, we  also adopted the pre-trained CNN, Places-CNN, as a descriptor generator. This CNN model is a multi-layer neural network that mainly consists of five convolutional layers, three max-pooling layers, and three fully connected layers. A max-pooling layer follows only the first, second, and fifth convolutional layers but not the third and fourth convolutional layers. The architecture is shown in Fig.\ref{figure1}. Given the remarkable performance of conv3 and pool5\cite{Hou-2015-p2238}, we selected these two patterned layers as representatives to combine with SeqSLAM in our subsequent comparative studies. These methods are respectively named as SeqCNNSLAM (conv3) and SeqCNNSLAM (pool5) correspond to conv3 and pool5 layers.

Unlike in SeqSLAM\cite{Milford-2012-p1643} that mainly involves three steps, in SeqCNNSLAM, the first two preprocessing steps of SeqSLAM are no longer applied. As illustrated in Algorithm \ref{algorithm3} (we only select one $ds$ and $V$ (trajectory speed) for illustration), SeqCNNSLAM uses the normalized output $X$ of conv3 and pool5 layers directly as the image descriptor. From steps 2 to 4, the difference matrix is obtained merely by directly calculating the Euclidean distance between the two layers. From steps 5 to 10, SeqCNNSLAM sweeps through the entire difference matrix to find the best matching sequence for the current frame. All experiments were based on OpenSeqSLAM implementation\cite{Niko-2013-p}. Default values were used for all parameters, except for sequence length ds (see Table \ref{table2}), which varied from 10 to 100. It should be noted that the first $ds/2$ images can not be matched, so the images' serial numbers $n$ should start from $ds/2+1$. But in order to make readers focus on the algorithm, we ignore this detail and set $n$ start from 1, which is default setting for all subsequent algorithms.

\begin{table}
	\caption{OpenSeqSLAM Parameters}

	\begin{tabular}{|c|c|c|}
		\hline
		Parameter & Value & Description\tabularnewline
		\hline
		$R{}_{windows}$ & 10 & Recent template range\tabularnewline
		\hline
		$V{}_{min}$ & 0.8 & Minimum trajectory speed\tabularnewline
		\hline
		$V{}_{max}$ & 1.2 & Maximum trajectory speed\tabularnewline
		\hline
		$ds$ & $\left\{ 10,20,\ldots,100\right\} $ & Sequence length\tabularnewline
		\hline
	\end{tabular}
	\label{table2}
\end{table}

\begin{figure*}
	\subfloat[]{\includegraphics[width=0.333\linewidth,height=4.3cm]{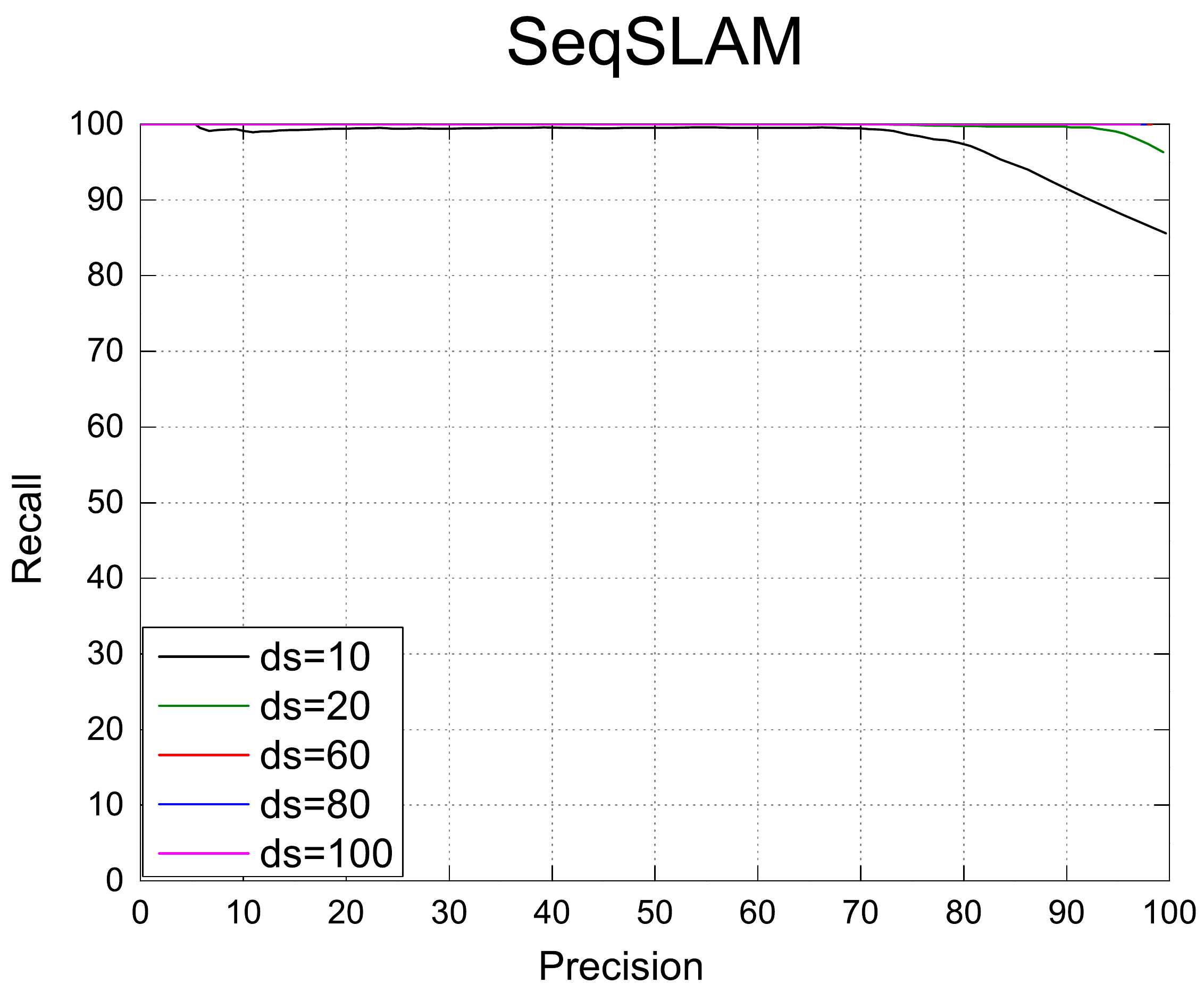}}\subfloat[]{\includegraphics[width=0.333\linewidth,height=4.3cm]{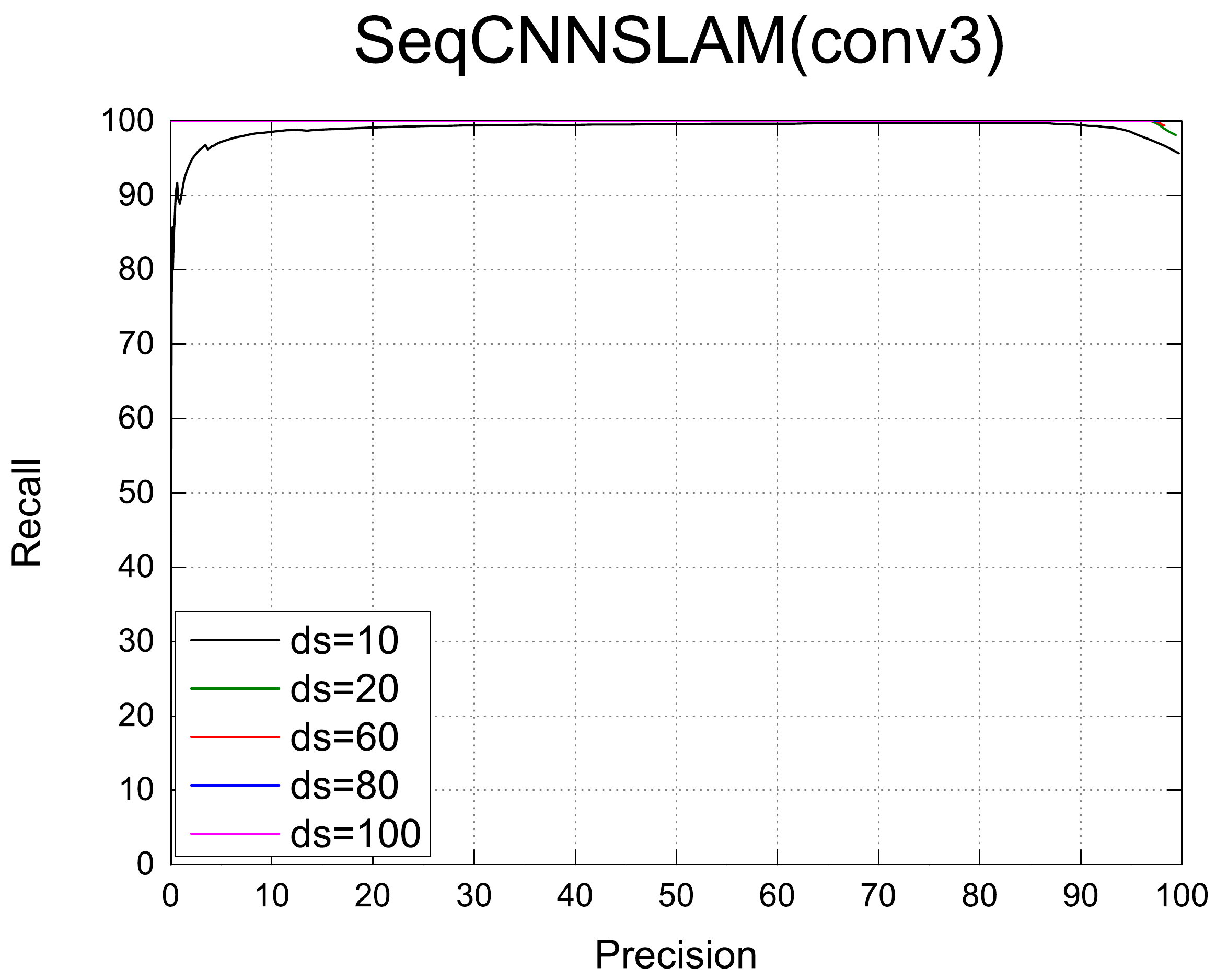}}\subfloat[]{\includegraphics[width=0.333\linewidth,height=4.3cm]{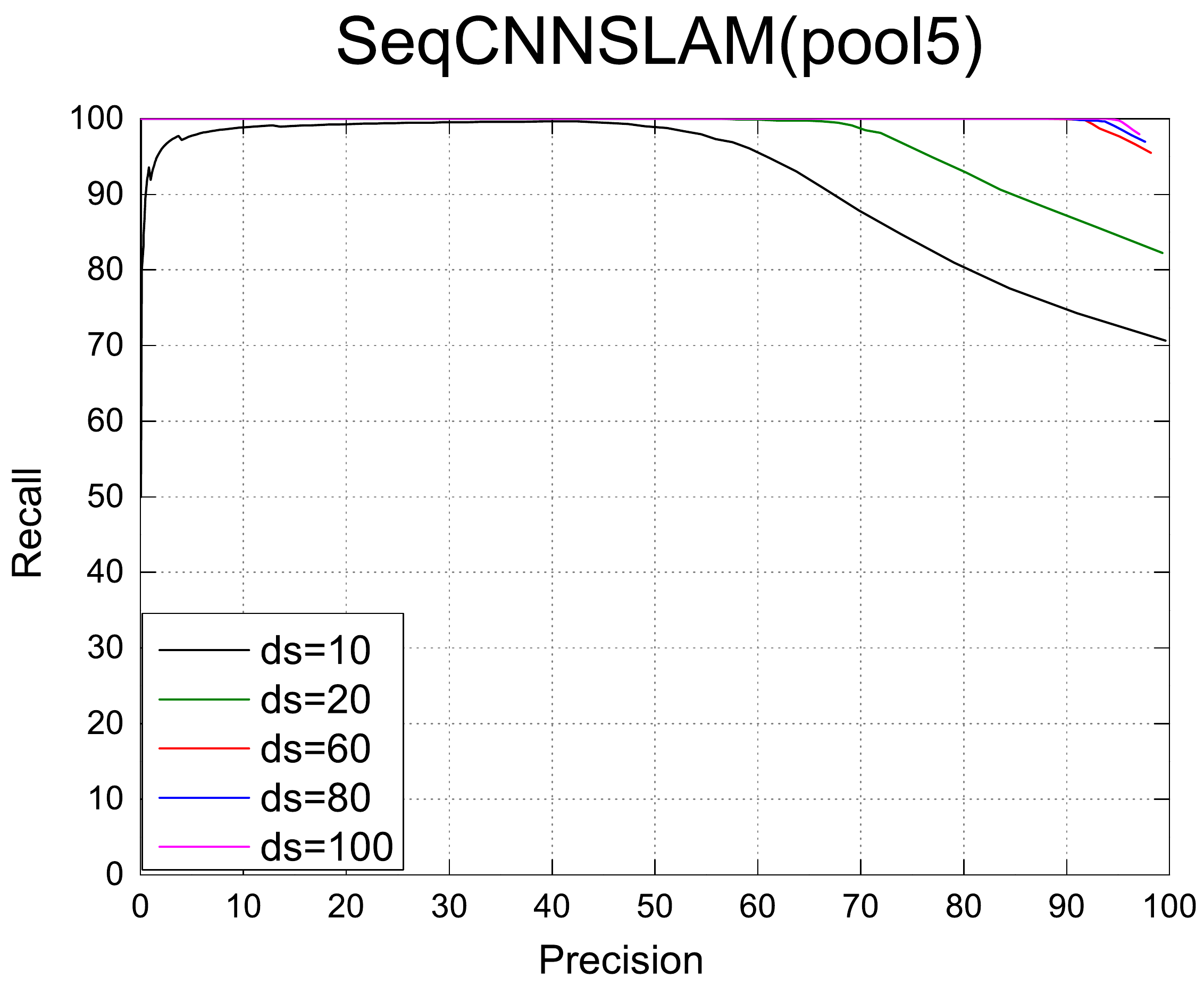}}
	
	\subfloat[]{\includegraphics[width=0.333\linewidth,height=4.3cm]{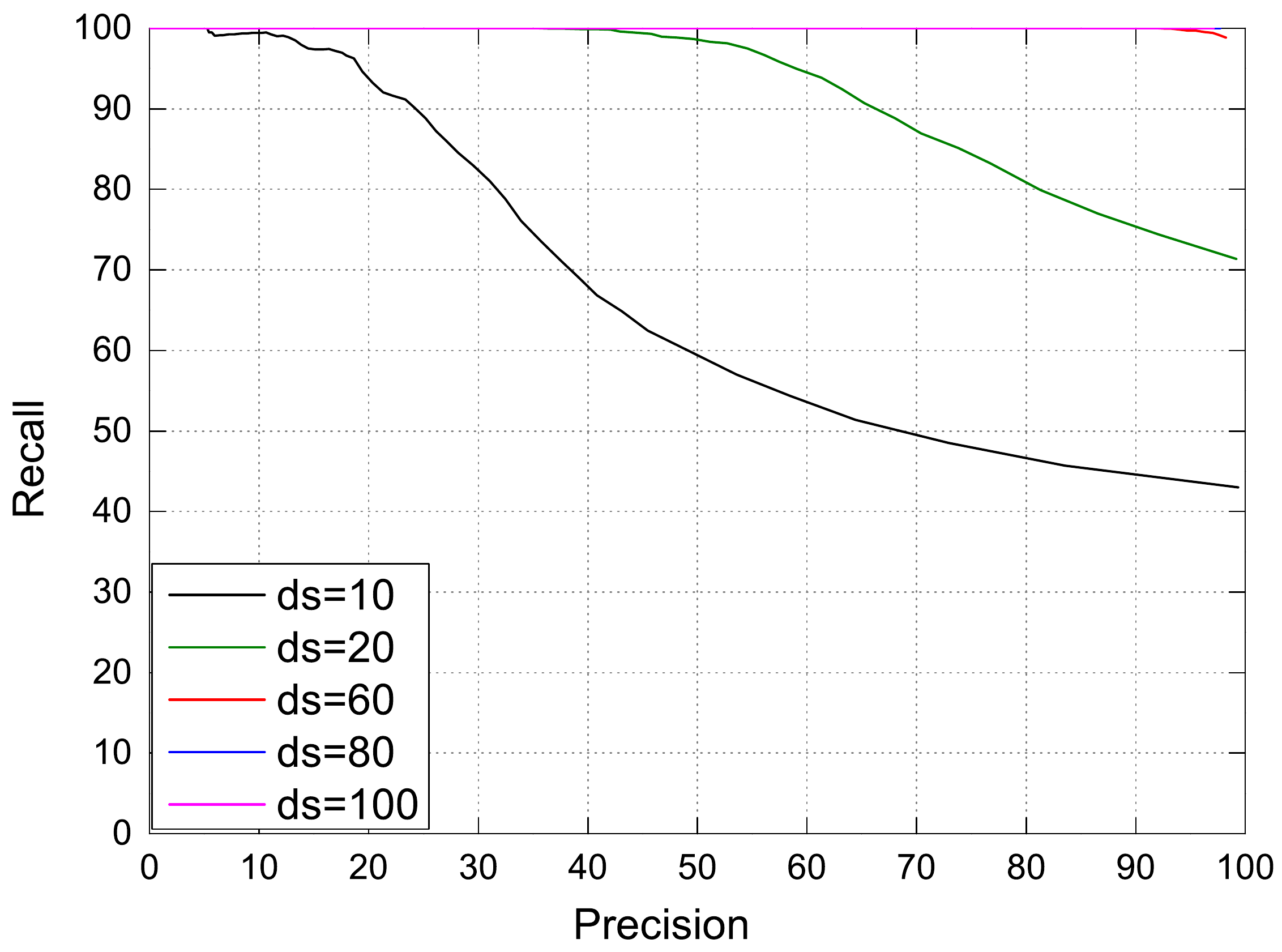}}\subfloat[]{\includegraphics[width=0.333\linewidth,height=4.3cm]{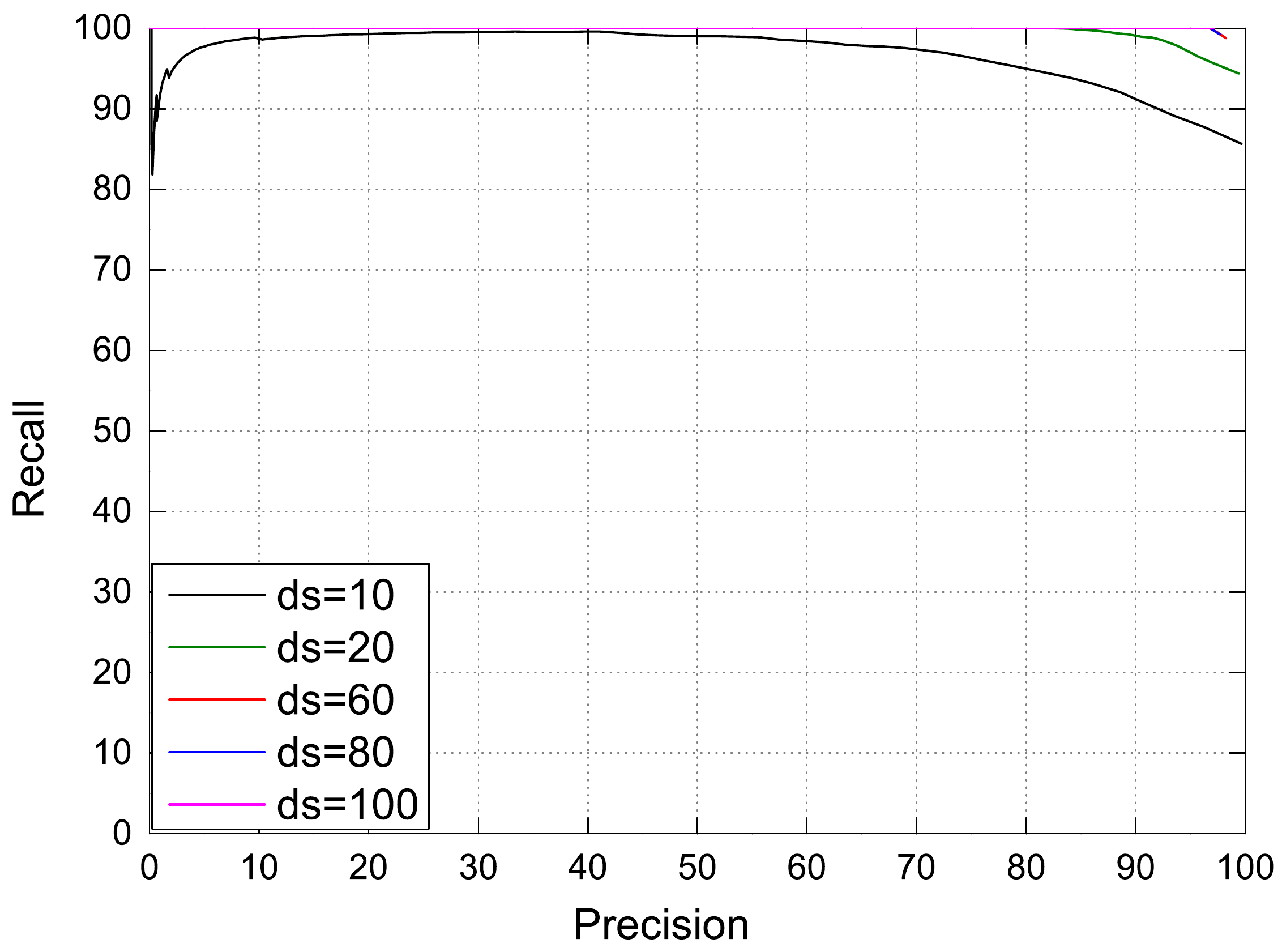}}\subfloat[]{\includegraphics[width=0.333\linewidth,height=4.3cm]{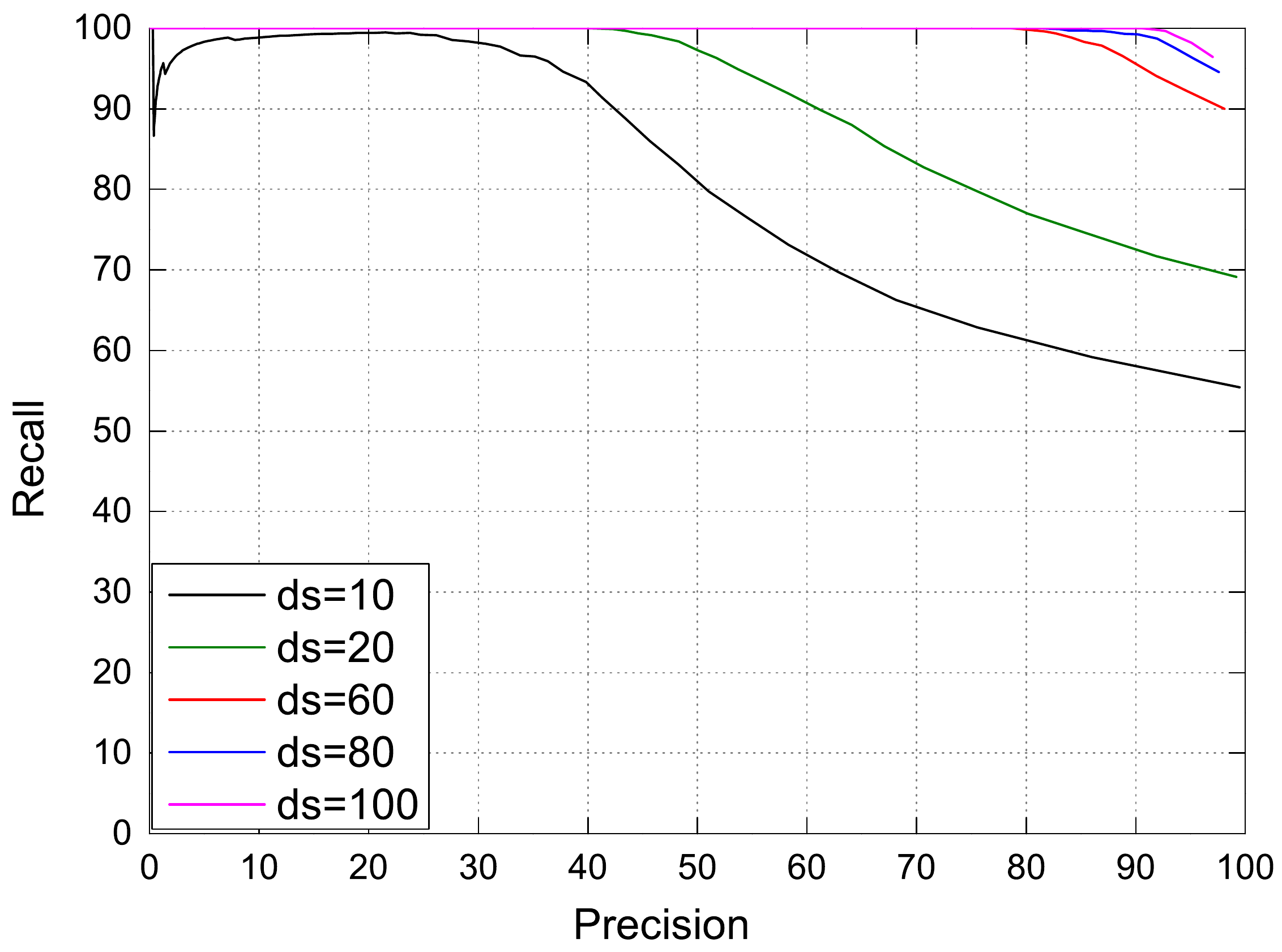}}
	
	\subfloat[]{\includegraphics[width=0.333\linewidth,height=4.3cm]{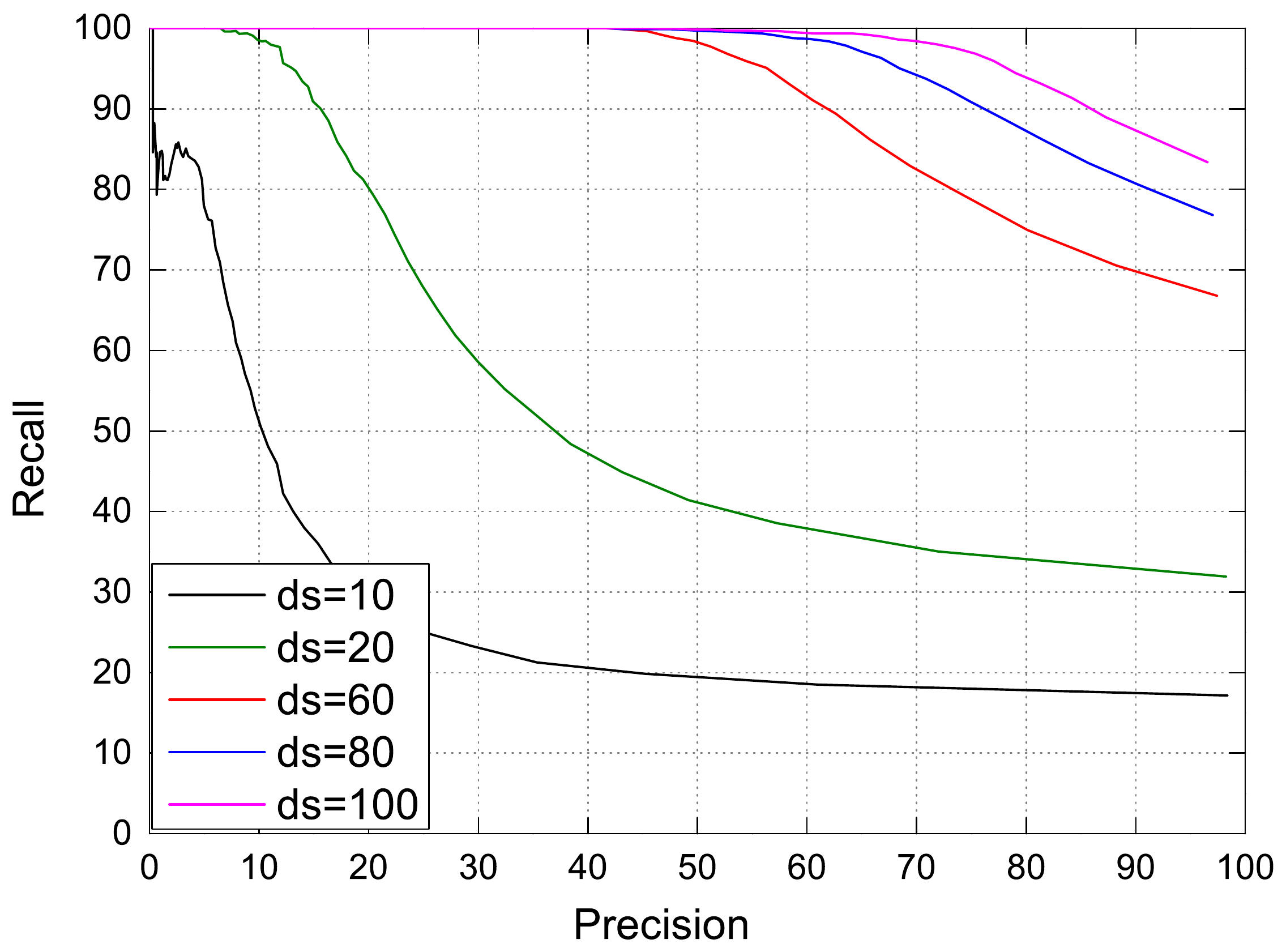}}\subfloat[]{\includegraphics[width=0.333\linewidth,height=4.3cm]{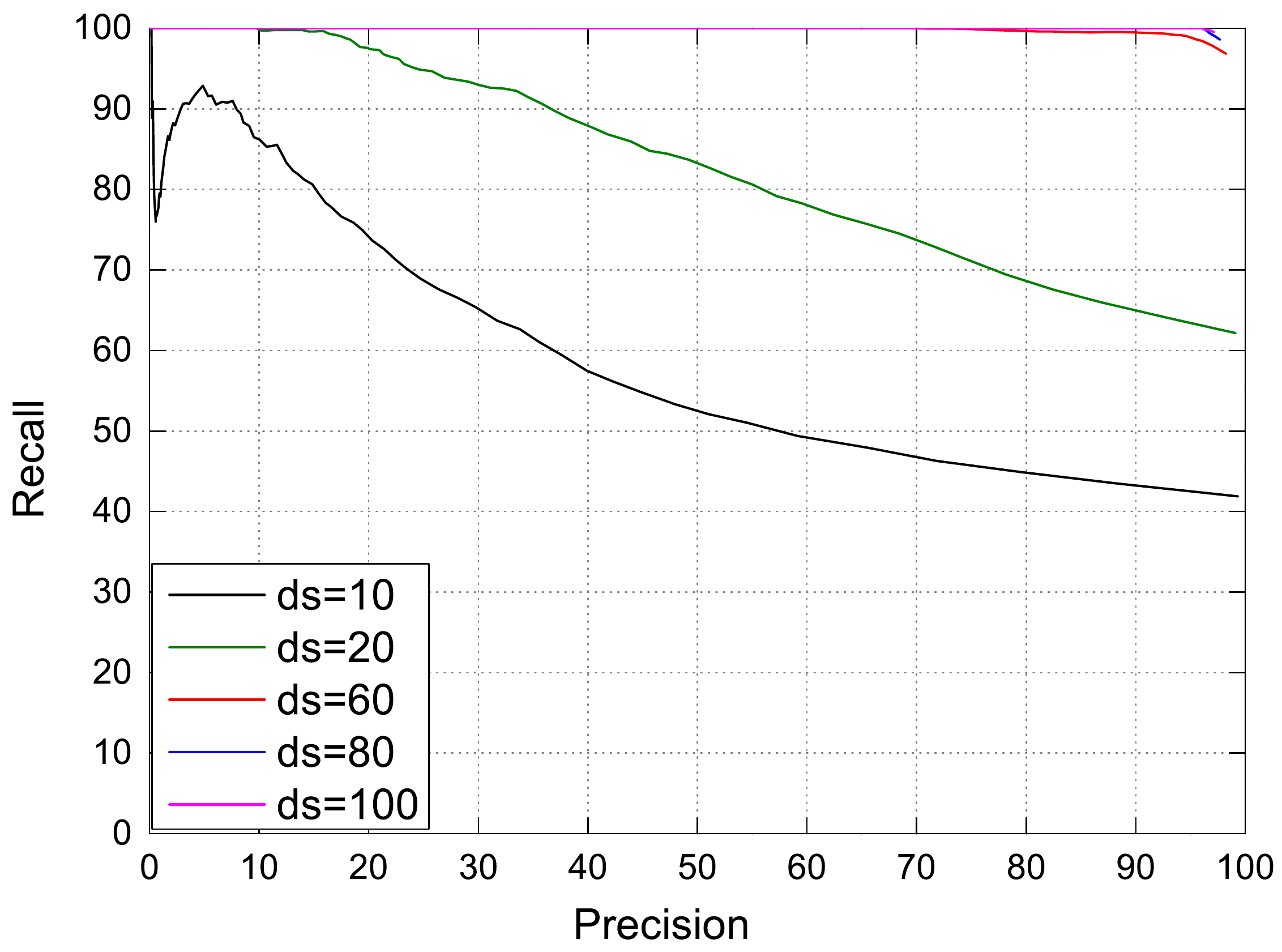}}\subfloat[]{\includegraphics[width=0.333\linewidth,height=4.3cm]{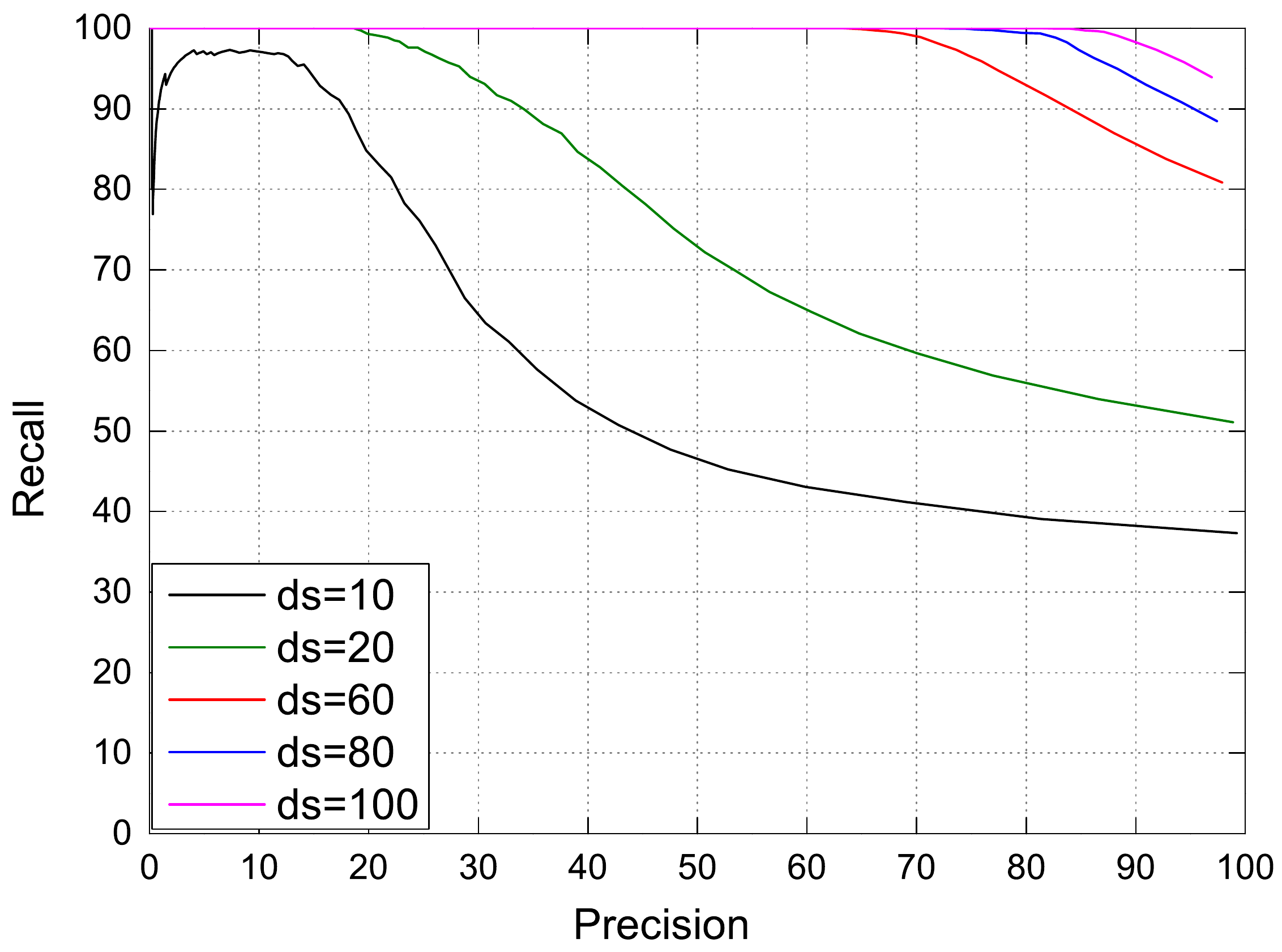}}
	
	\subfloat[]{\includegraphics[width=0.333\linewidth,height=4.3cm]{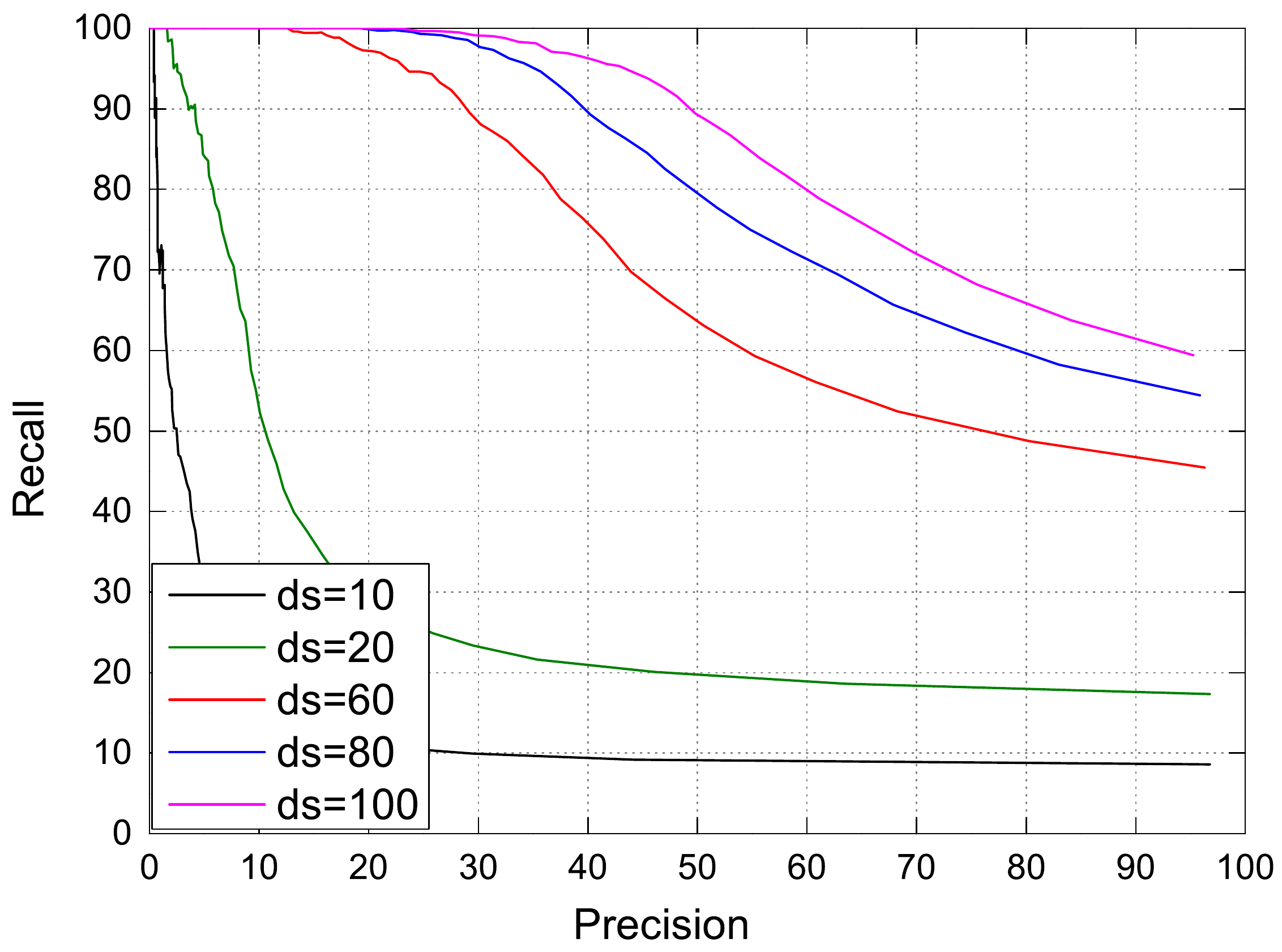}}\subfloat[]{\includegraphics[width=0.333\linewidth,height=4.3cm]{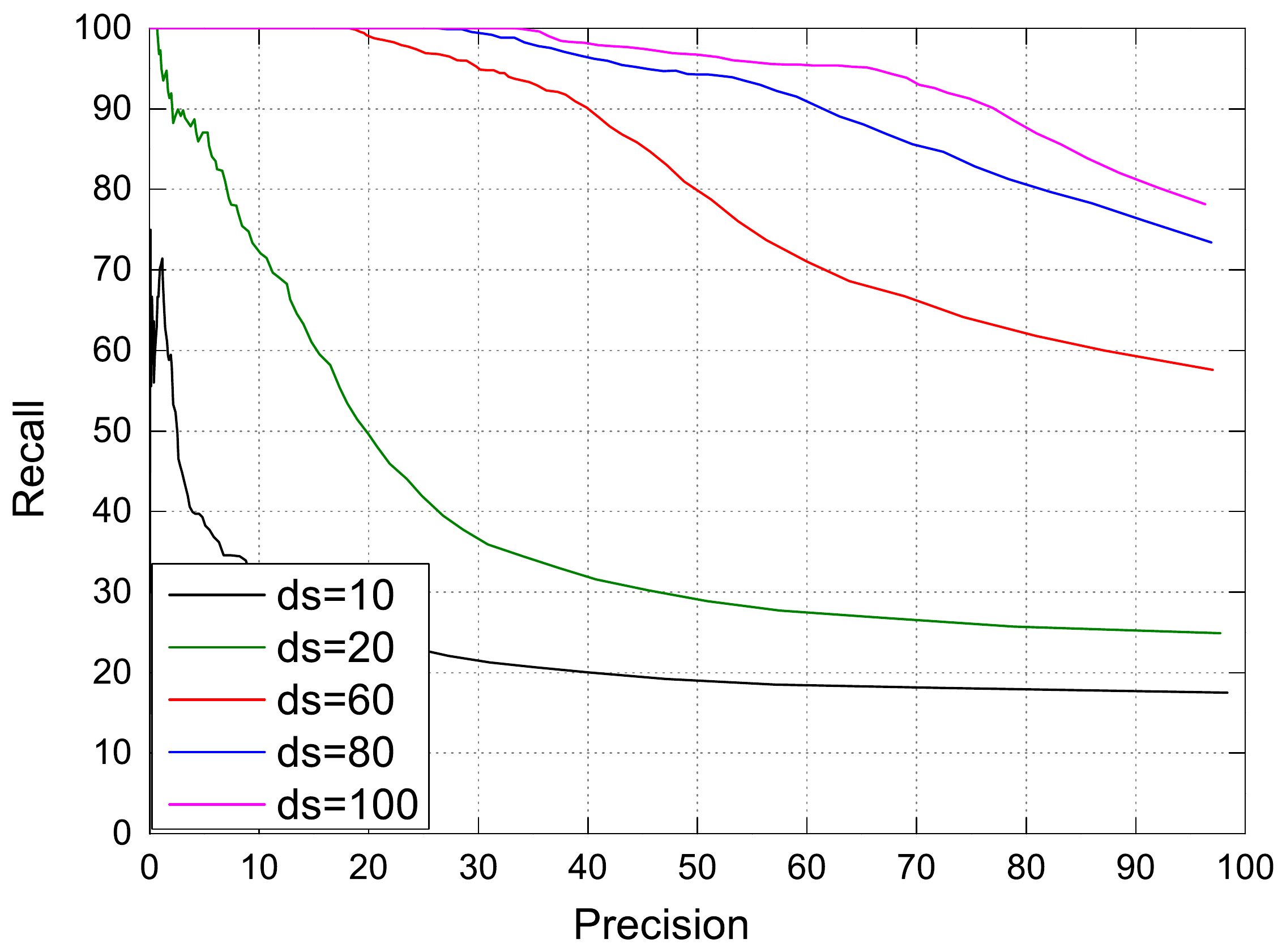}}\subfloat[]{\includegraphics[width=0.333\linewidth,height=4.3cm]{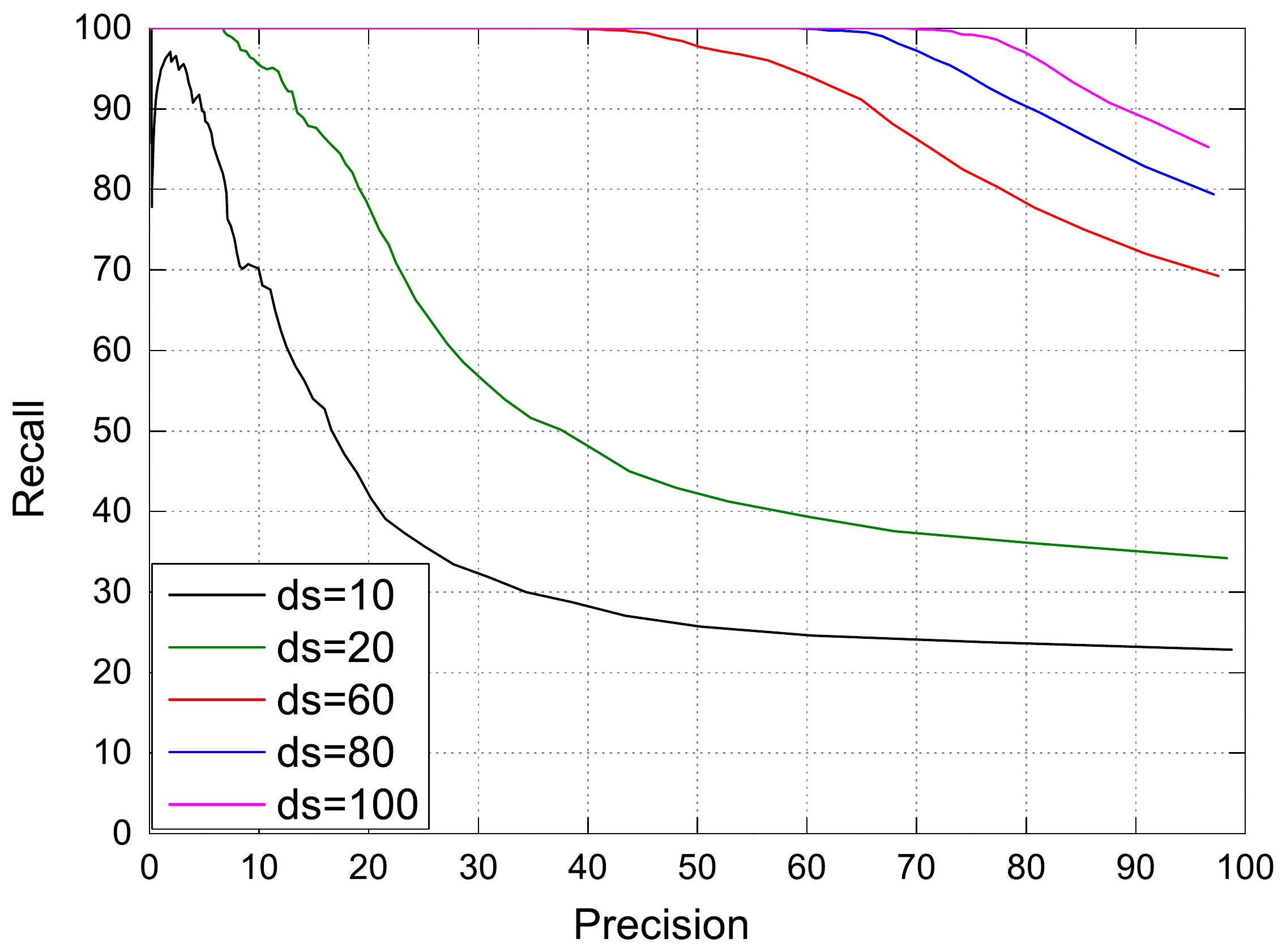}}
	
	\caption{Performance of SeqSLAM, SeqCNNSLAM(conv3) and SeqCNNSLAM(pool5) in
		the Nordland dataset with only condition change (top line) as well
		as with both condition and viewpoint change by 4.2\%, 8.3\%, and 12.5\%
		shift (the second, third and bottom lines). SeqSLAM(left line) and
		SeqCNNSLAM(conv3) (middle line) achieve comparable performance on
		only condition change (0\% shift), and out-performs SeqCNNSLAM(pool5)
		(right line). However, with the increment of in the viewpoint, SeqCNNSLAM(pool5)
		presents a more robust performance than SeqSLAM and SeqCNNSLAM(conv3),
		especially at 12.5\% shift.}
	\label{figure5}
	
\end{figure*}

\begin{figure}
	\subfloat{\includegraphics[width=0.25\columnwidth]{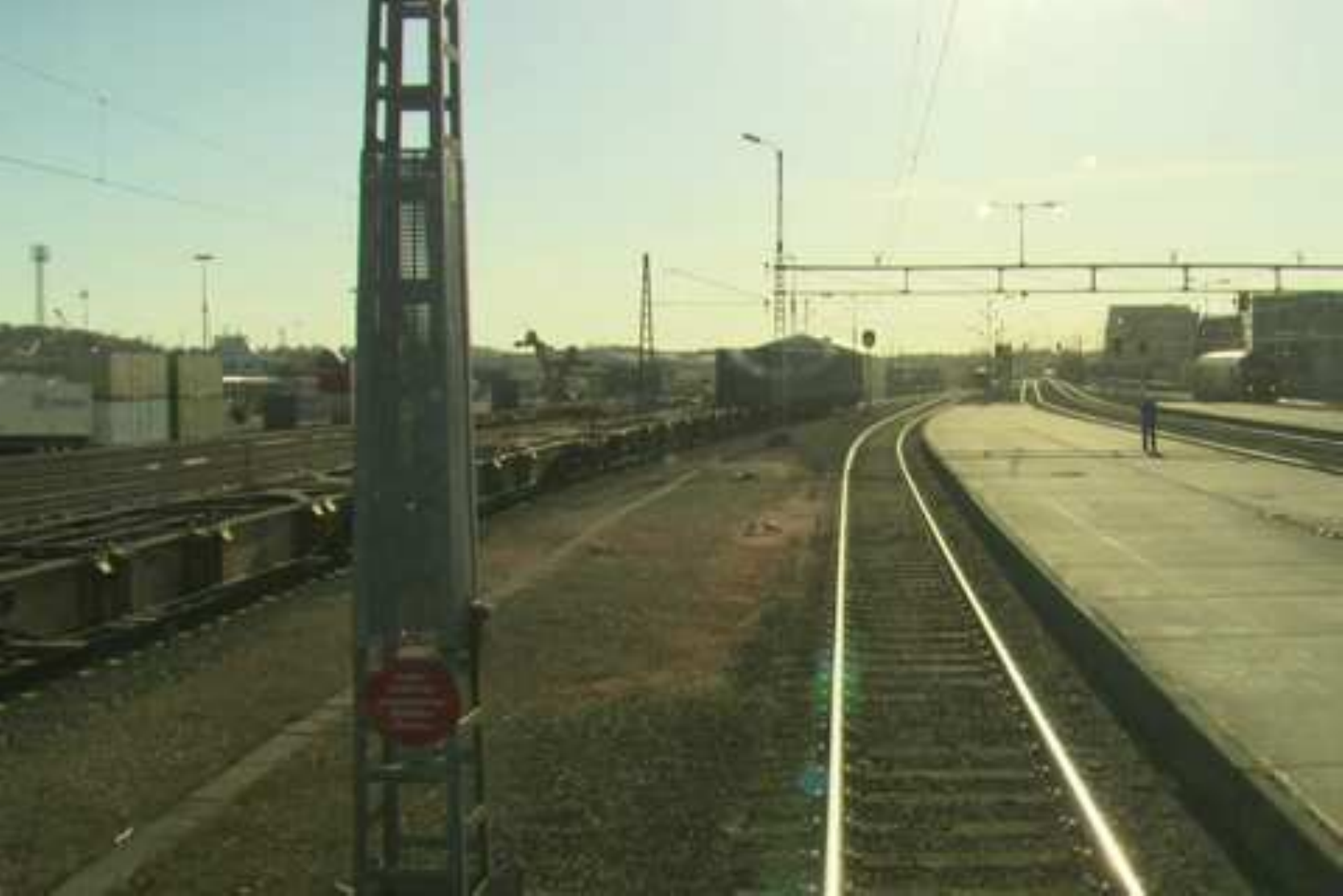}}\subfloat{\includegraphics[width=0.25\columnwidth]{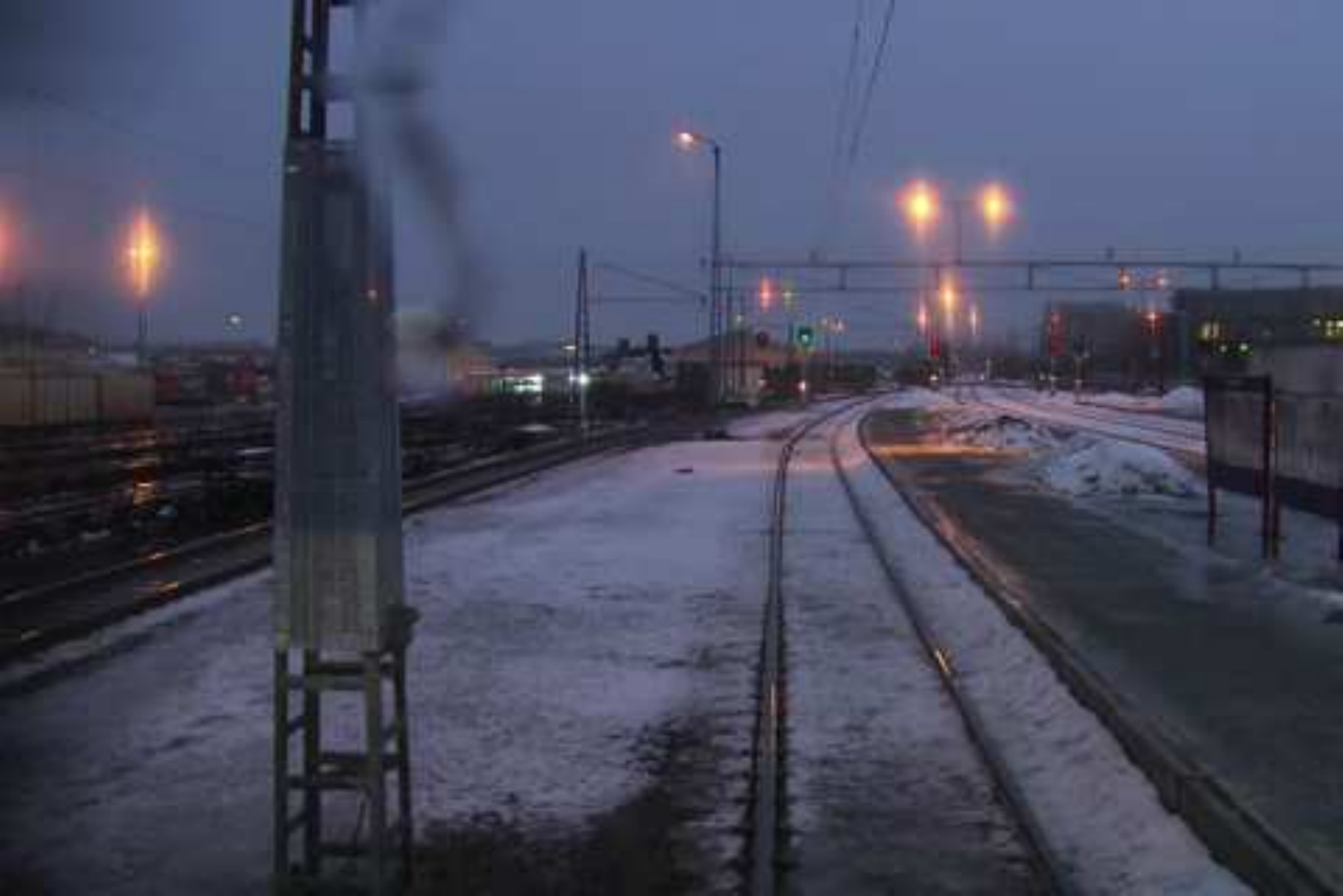}}\subfloat{\includegraphics[width=0.25\columnwidth]{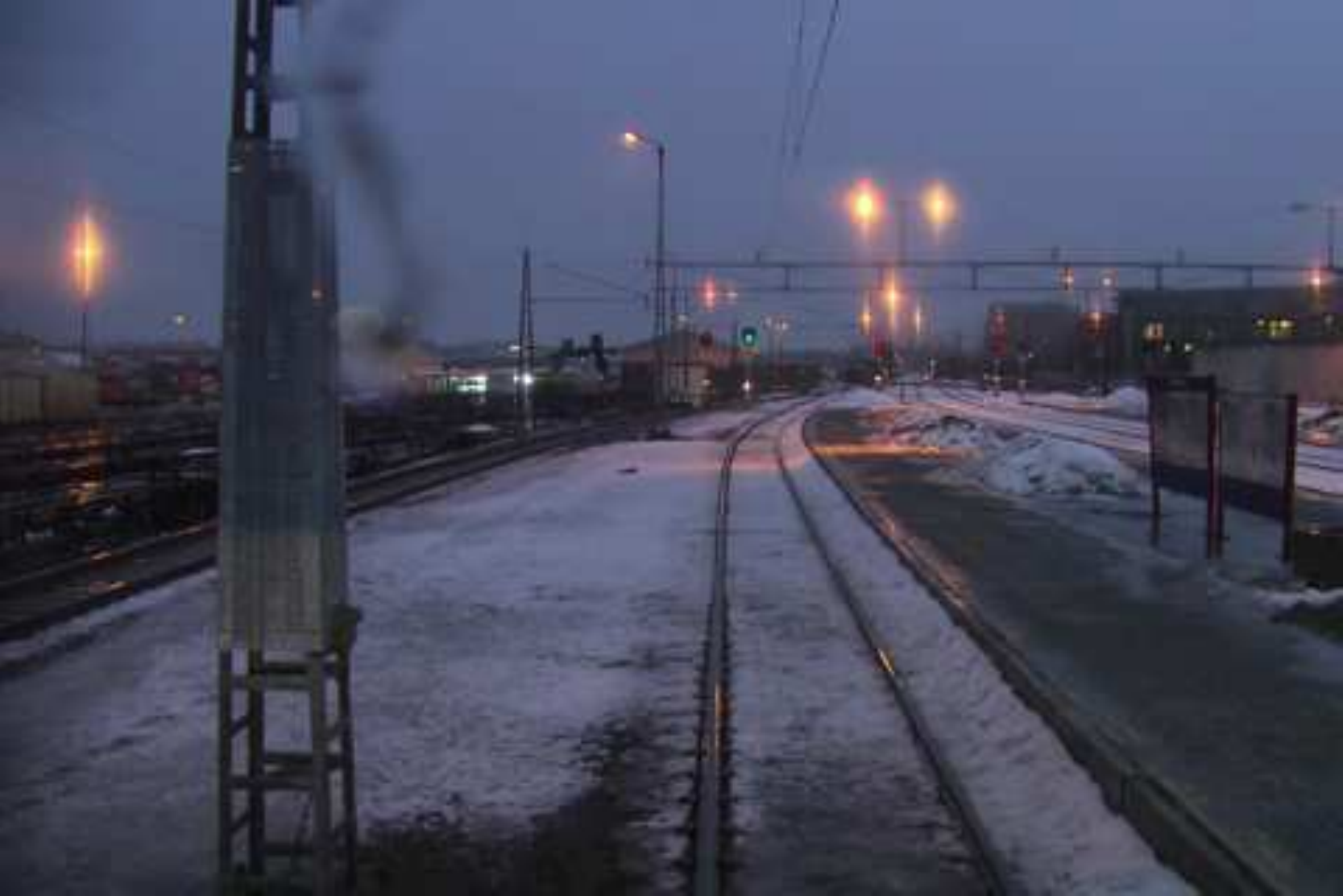}}\subfloat{\includegraphics[width=0.25\columnwidth]{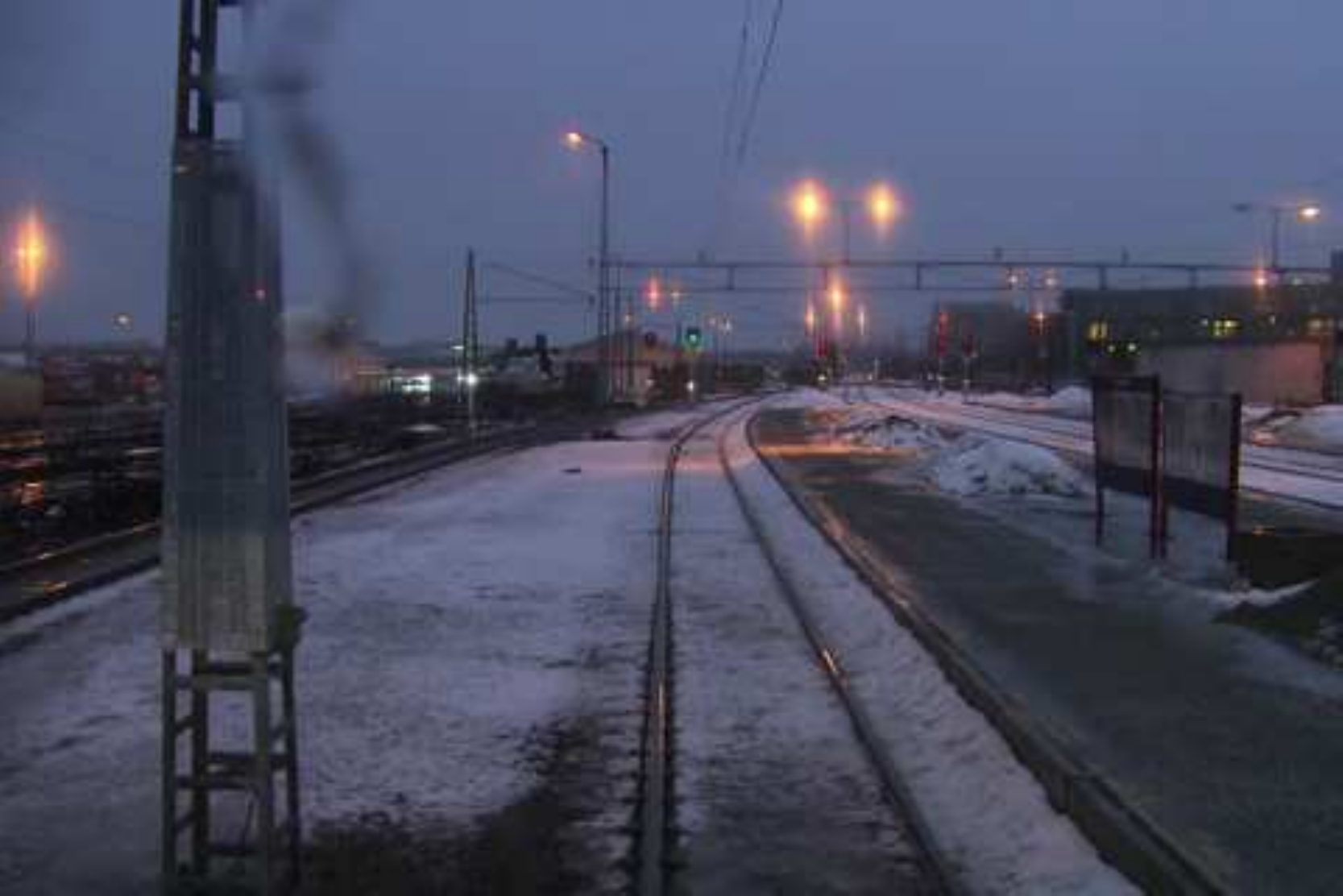}}
	
	\subfloat{\includegraphics[width=0.25\columnwidth]{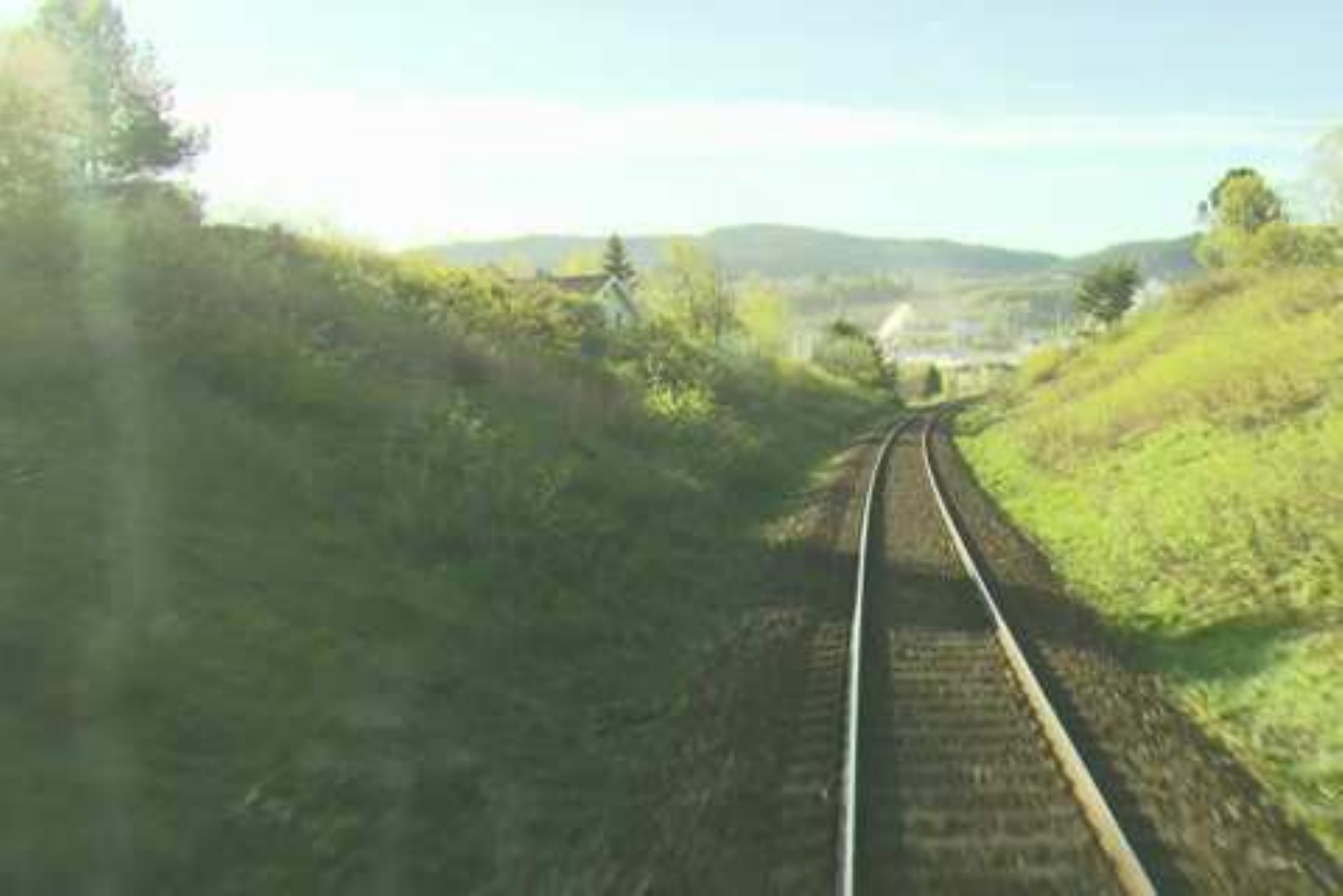}}\subfloat{\includegraphics[width=0.25\columnwidth]{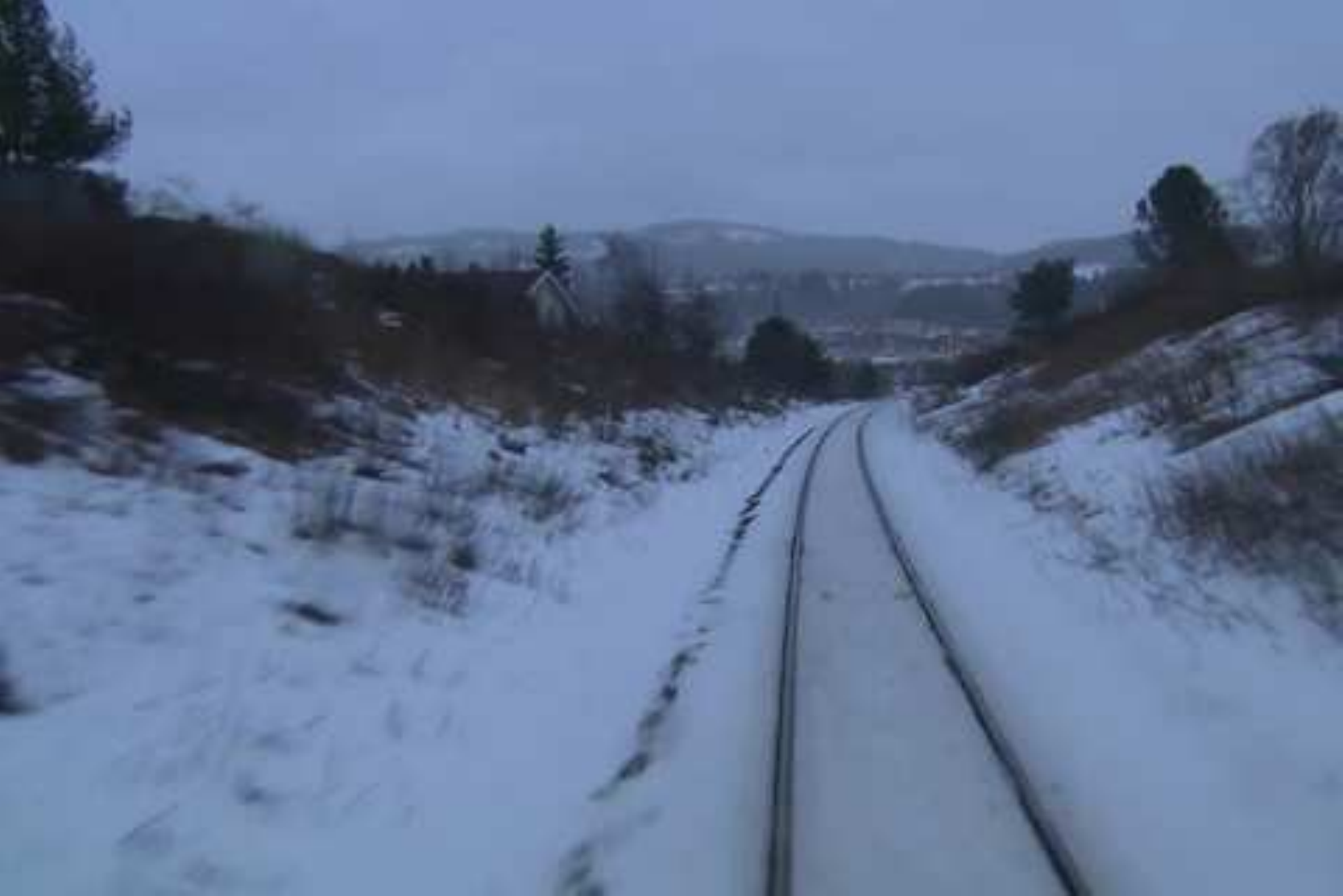}}\subfloat{\includegraphics[width=0.25\columnwidth]{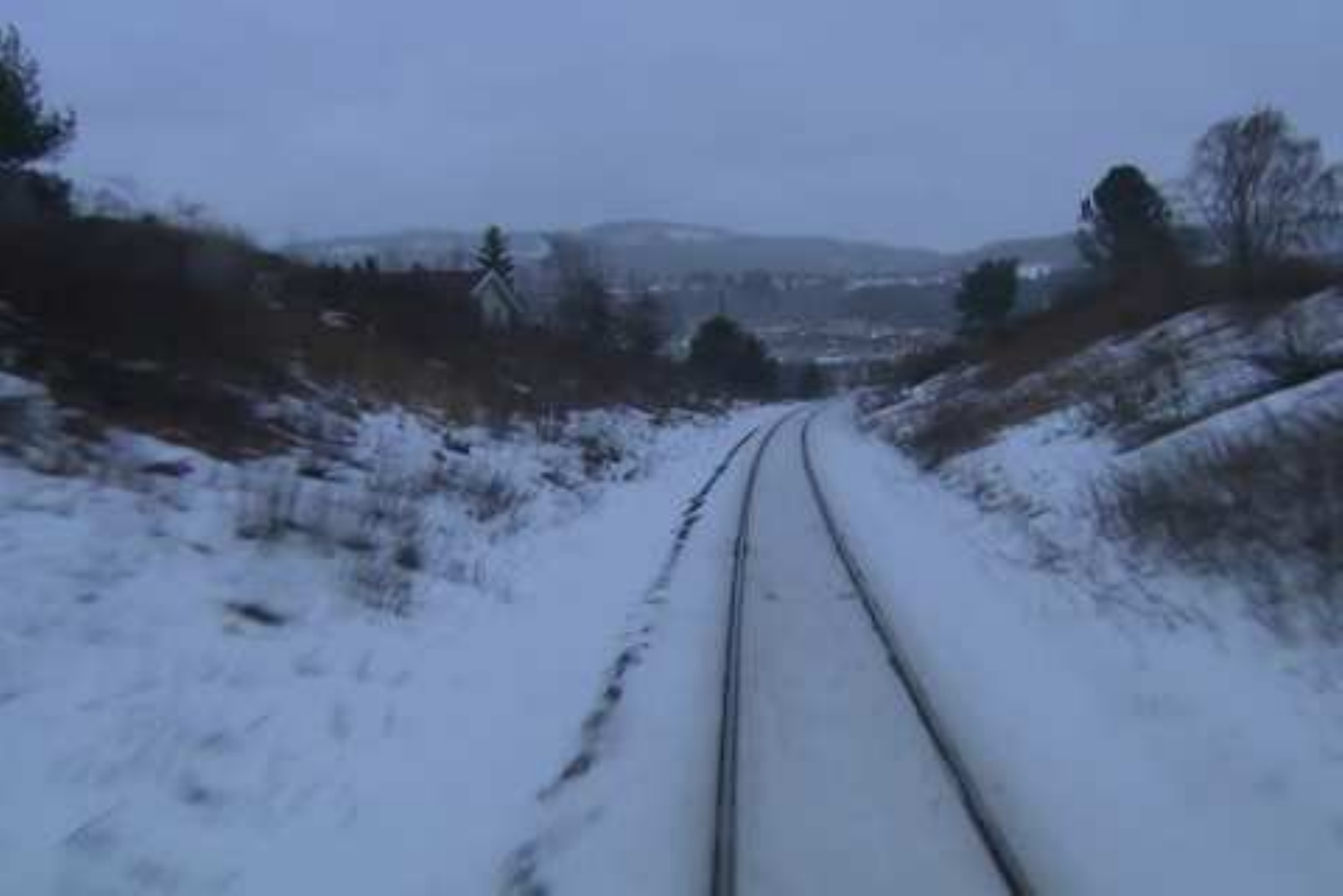}}\subfloat{\includegraphics[width=0.25\columnwidth]{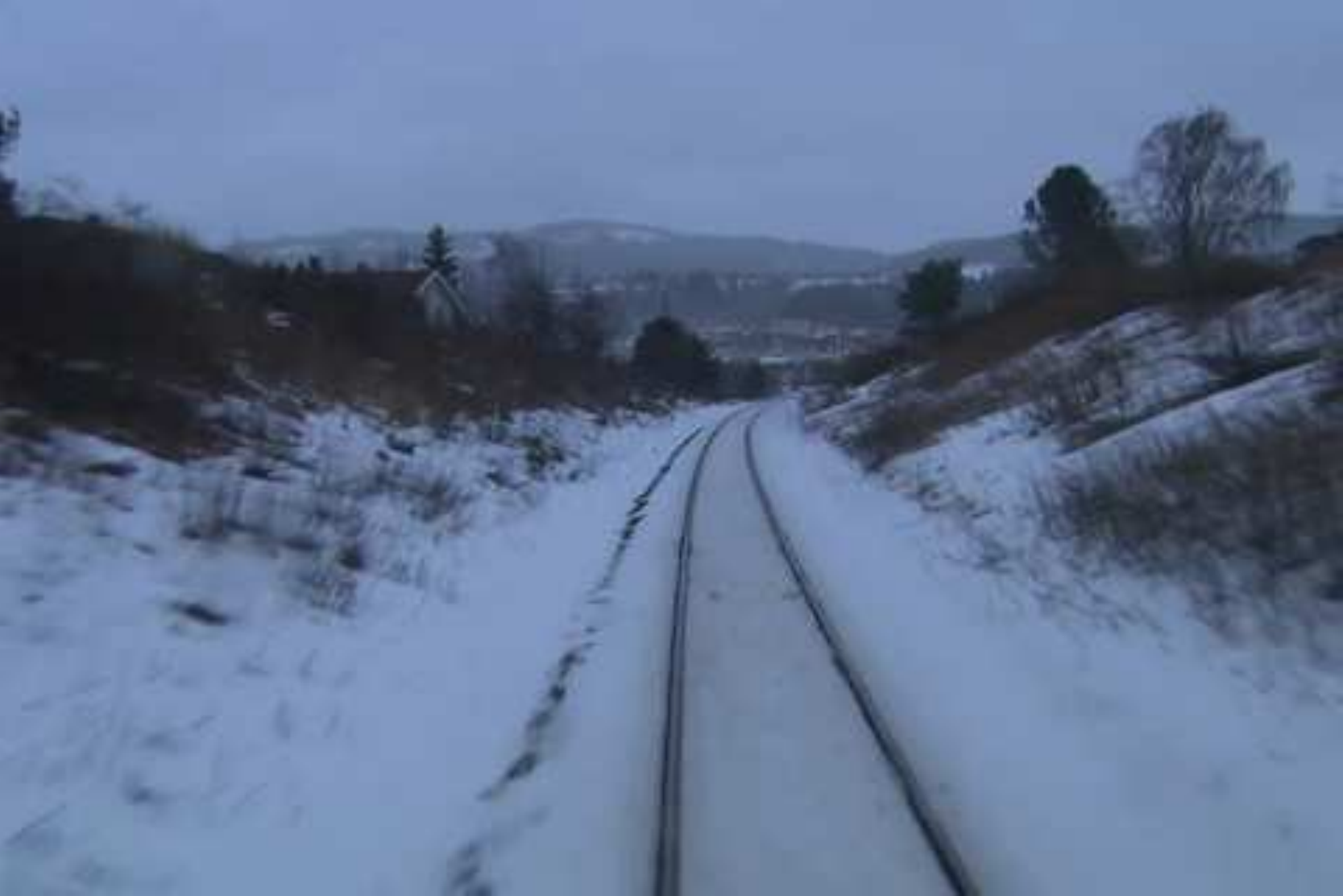}}
	
	\caption{Cropped image samples in Nordland. Each row of images is cropped from
		the same image. The left to right columns represent spring 0\%, winter
		4.2\%, winter 8.3\%, and winter 12.5\%.}
	\label{figure4}
\end{figure}

To evaluate our LCD algorithm's actual performance, we thoroughly investigate SeqCNNSLAM's robustness against viewpoint and condition changes. To experiment on the robustness of the algorithm against viewpoint change, we crop the 640*320 down-sampled images of the spring and winter datasets to 480*320 pixels. For the winter images, we simulate viewpoint change by shifting these down-sampled crops to 4.2\%, 8.3\%, and 12.5\% denoted as winter 4.2\%, winter 8.3\% and winter 12.5\%, respectively. The shift pixels of winter images are 20, 40, and 60 pixels. Meanwhile, the cropped images of the spring dataset are called spring 0\%, as illustrated in Fig.\ref{figure4}. To compare with SeqSLAM, we also cropped the 64*32 pixel down-sampled images of SeqSLAM to 48*32 pixels and shifted the images at an equal proportion as in SeqCNNSLAM, resulting in 2, 4, and 6 pixels shift.

\begin{algorithm}
	\caption{SeqCNNSLAM}

	\textbf{Require:} $S=\left\{ \left(x_{i},y_{i}\right)\right\} _{i=1}^{N}$:
	dataset for LCD containing $N$ images, $x_{i}$: input images, $y_{i}$
	the ground truth of matching image's serial number of $x_{i}$; $\left\{ X_{i}\right\} _{i=1}^{N}$:
	CNN descriptors of $x_{i}$; $ds$: sequence length; $D\in R^{N\times N}$:
	difference matrix, $D_{ij}$ : a element of $D$; $Dis(X,Y)$: Euclidean
	distance between X and Y; $V$: trajectory speed.
	
	\textbf{Ensure:}$\tilde{y_{n}}$: the matching image's serial number
	of $x_{n}$ determined by LCD algorithm.
	
	01$\quad$Initial: $i=1$, $j=1$
	
	02$\quad$for $i=1:N$, $j=1:N$:
	
	03$\quad$$\quad$$\quad$$D_{ij}=Dis(X,Y)$
	
	04$\quad$end for
	
	05$\quad$for $n=1:N$
	
	07$\quad$$\quad$$\quad$ for $q=1:N$
	
	08$\quad$$\quad$$\quad$$\quad$$\quad$$sum_{nq}=$Cal-Seq-Dif$(n,q,ds)$;
	
	09$\quad$$\quad$$\quad$end for
	
	10$\quad$$\quad$$\quad$$\tilde{y}_{n}=min(sum_{n})$
	
	11$\quad$end for
	\label{algorithm3}
\end{algorithm}

We conduct these experiments on the following set of datasets.
\begin{itemize}
	\item Nordland: Spring and winter datasets are used to form a pair with only condition change as the baseline of the Nordland dataset.
	\item Nordland: Spring 0\% are used to form pairs with winter 4.2\%, winter 8.3\% or winter 12.5\%  to construct a set of datasets with both viewpoint and condition change.
	\item Gardens Point: Day-left and night-right are used to form a pair with both condition change and viewpoint change.
\end{itemize}

For comparison, we also conducted several tests on SeqSLAM in the first two set of datasets and on Change Removal in the last dataset.

Fig.\ref{figure5} shows the resulting precision-recall curves in Nordland dataset, and Table \ref{table3} shows the runtime for each method with the variation in $ds$.

Fig.\ref{figure5}(a) and Fig.\ref{figure5}(c) show that SeqSLAM\cite{Milford-2012-p1643} and SeqCNNSLAM (conv3) exhibit comparable performance against condition changes only and present slightly better robustness than SeqCNNSLAM (pool5) when $ds$ is set to the same value. However, with the increment in viewpoint, SeqCNNSLAM (pool5) achieves overwhelming performance compared with SeqSLAM and SeqCNNSLAM (conv3), as illustrated in Fig.\ref{figure5}(d) to Fig.\ref{figure5}(l), especially at 12.5\% shift.

\begin{figure}
	\subfloat{\includegraphics[width=1\columnwidth]{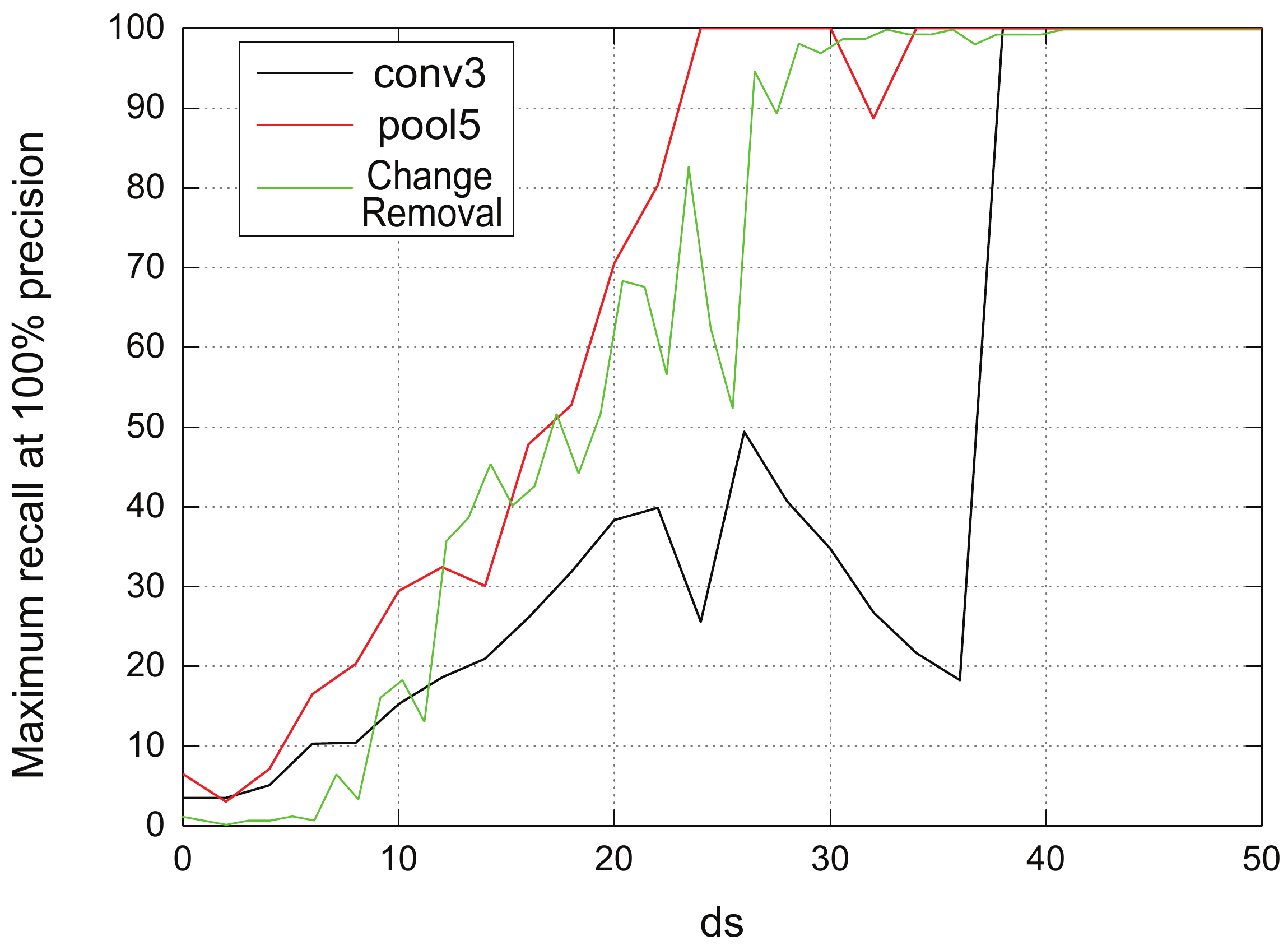}}
	
	\caption{Performance of SeqCNNSLAM and Change Removal in the Gardens Point
		dataset with both viewpoint and condition change.}
	
	\label{figure6}
\end{figure}

Fig.\ref{figure6} shows the best performance of Change Removal\cite{Lowry-2015-p} (green line) on the day-left and night-right parts of the Gardens Point dataset. Evidently, SeqCNNSLAM(pool5) achieves better performance than Change Removal\cite{Lowry-2015-p}. Because when $ds$ is set to the same value, SeqCNNSLAM(pool5) achieves better recall when the precision is maintained at 100\%.

From these experiments, we conclude that SeqSLAM\cite{Milford-2012-p1643} and SeqCNNSLAM (conv3) are suitable for dealing with scenes of severe condition change but minor viewpoint change. When a scene contains severe condition and viewpoint change, SeqCNNSLAM (pool5) is the more sensible choice compared with the other methods.

\section{Approaches to Realize Real-Time SeqCNNSLAM}
Besides viewpoint and condition invariance,  real-time performance is another important performance metric for LCD.  For SeqCNNSLAM, the time to calculate the difference matrix is the key limiting factor for large-scale scene, as its runtime is proportional to the square of the number of images in the dataset. In this section, we provide an efficient acceleration method of SeqCNNSLAM (A-SeqCNNSLAM) by reducing the number of candidate matching images for each image. Then, we present an online method to select the parameters of A-SeqCNNSLAM to enable its applicability for unseen scenes.
\subsection{A-SeqCNNSLAM: Acceleration method of SeqCNNSLAM}
SeqCNNSLAM (pool5) shows a satisfactory performance when facing viewpoint and condition change simultaneously, but the runtime of this method (illustrated in Table \ref{table3}) is too long. The reason for this phenomenon is that for any image in the dataset, SeqCNNSLAM (pool5) performs a literal sweep through the entire difference matrix to find the best matching sequence, similar to SeqSLAM\cite{Milford-2012-p1643}. To obtain the entire difference matrix, the computer needs to calculate the Euclidean distance between any two images' CNN descriptors in the dataset. So if the dataset contains $N$ images, the time complexity of obtaining the difference matrix is proportional to the square of $N$. Furthermore, the LCD algorithm must perform $N$ searches to find the matching image for any image. Evidently, with the increase in the number of images, the increasing  rate of overhead of SeqCNNSLAM(pool5) is formidable. Hence, directly using the SeqCNNSLAM(pool5) may not be suitable to deal with the large-scale scene.

\begin{table*}
\caption{Runtime of SeqCNNSLAM}
\begin{centering}
\begin{tabular}{|c|c|c|c|c|c|c|}
	\hline
	Algorithms & \backslashbox{Viewpoint}{ds} & 10  & 20 & 60 & 80 & 100\tabularnewline
	\hline
	\multirow{4}{*}{SeqCNNSLAM(conv3)} & 0 & 4111.034s &  4131.820s & 3988.761s & 4035.855s & 4051.425s\tabularnewline
	\cline{2-7}
	& 20 & 4161.219s &  4155.817s  & 4155.309s & 4163.036s & 4160.891s\tabularnewline
	\cline{2-7}
	& 40 & 4153.573s & 4159.118s &  4163.875s & 4164.591s & 4164.014s\tabularnewline
	\cline{2-7}
	& 60 & 4160.501s & 4154.367s &  4162.638s & 4157.301s &  4162.374s\tabularnewline
	\hline
	\multirow{4}{*}{SeqCNNSLAM(pool5)} & 0 & 626.554s & 641.893s &  649.681s &  653.731s & 654.196s\tabularnewline
	\cline{2-7}
	& 20 & 640.814s  & 641.955s & 644.552s & 646.598s & 647.785s\tabularnewline
	\cline{2-7}
	& 40 & 640.289s  & 641.947s & 644.278s & 645.876s & 646.537s\tabularnewline
	\cline{2-7}
	& 60 & 640.013s  & 641.483s & 644.072s & 646.107s & 646.798s\tabularnewline
	\hline
\end{tabular}
\label{table3}
\par\end{centering}
\end{table*}

The movement of a robot is continuous in both time and space, hence the adjacently collected images are of a high similarity. Therefore, we may infer that if an image A is matched with another image B, A's adjacent image is also very likely to find its matching images in the adjacent range of image B.

Given the relationship between adjacent images, we greatly accelerate the execution of SeqCNNSLAM(pool5) by reducing the number of candidate matching images. For a certain image, we first choose $K$ images as its \emph{reference images}. Then, we select a collection of $Num$ images for each reference images, where $Num$ is denoted as the size of the  matching range. In this way, we may get at most $(K(Num+1))$ candidate matching images, considering the matching ranges of different reference images may overlap. These candidate matching images are corresponding to images' flag value in Algorithm \ref{algorithm4}.

\begin{figure}
	\subfloat{\includegraphics[width=1\columnwidth]{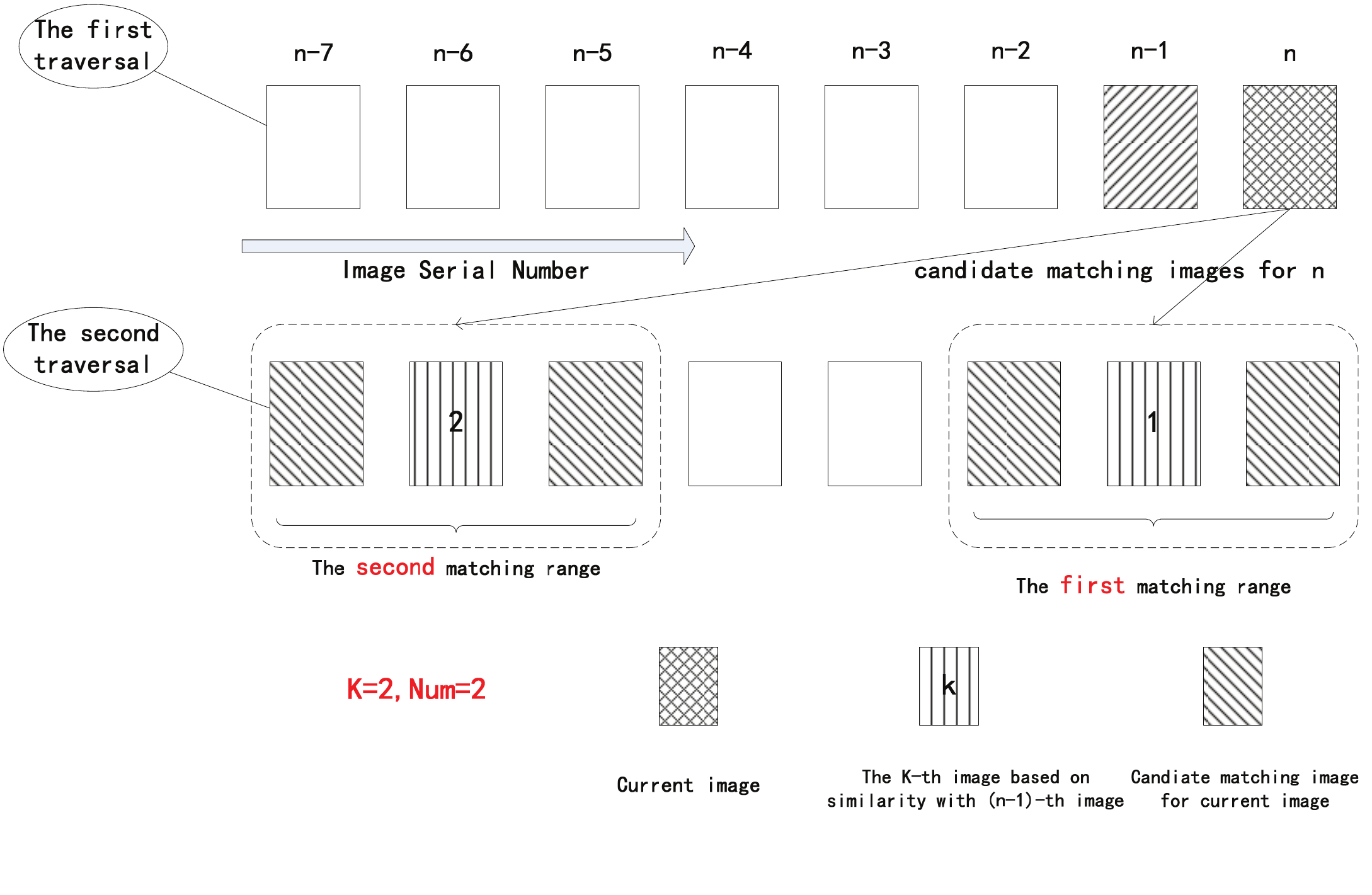}}
	
	\caption{Example of candidate matching ranges for the $n$-th image $K$ and $Num$
		are set to 2.}
	\label{figure7}
	
\end{figure}

To be more specific, for the $n$-th image, its reference images are the first $K$ images which are of the shortest distance with to $(n-1)$-th image. By setting the $k$-th reference image as the middle point, as shown in Fig.\ref{figure7}, we choose $(Num/2)$ images on both sides of the reference image and construct the $k$-th matching range. For example, when we set $K=2$ and $Num=2$, as shown in Fig.\ref{figure7}, the location of candidate matching ranges for the current image depends only on the first two images' serial number that is most similar to that of the $(n-1)$-th image.

As illustrated in Algorithm \ref{algorithm4}, from steps 2 to 4, A-SeqCNNSLAM is initialized by calculating the first $(ds+1)$ columns of the difference matrix and sweeping through these columns to determine the best matching sequence of the first image. Meanwhile, the first $K$ images that are most similar to it are recorded, and these $K$ images are set as the middle image of $K$ matching ranges containing $Num$ images for the next image by setting their flags to 1, as shown in steps 6 to 16, while $n$ is equal to 1.

\begin{algorithm}
	\caption{A-SeqCNNSLAM}

	\textbf{Require:} $S=\left\{ \left(x_{i},y_{i}\right)\right\} _{i=1}^{N}$:
	dataset for LCD containing $N$ images, $x_{i}$: input images, $y_{i}$:
	the ground truth of matching image's serial number of $x_{i}$; $\left\{ X_{i}\right\} _{i=1}^{N}$:
	CNN descriptors of $x_{i}$; $ds$: sequence length; $D\in R^{N\times N}$:
	difference matrix, $D_{ij}$: a element of $D$; $Dis(X,Y)$: Euclidean
	distance between X and Y; $K$: the number of matching ranges for
	an image; $k$: serial number of a matching range; $Num$: the lengths
	of each matching sequence; $sort(Y)$: sort vector $Y$ from small
	to large; $X.flag$: flag value of CNN descriptor $X$; $V$: trajectory
	speed.
	
	\textbf{Ensure:} $\tilde{y_{n}}$: the matching image's serial number
	of $x_{n}$ determined by LCD algorithm.
	
	01$\quad$Initial: $i=1$, $j=1$, $X.flag=1$;
	
	02$\quad$for $i=1:N$, $j=1:ds+1$
	
	03$\quad$$\quad$$\quad$$D_{ij}=Dis(X_{i},X_{j})$
	
	04$\quad$end for
	
	05$\quad$for $n=1:N$
	
	06$\quad$$\quad$$\quad$for $q=1:N$ and $X_{q}.flag==1$
	
	07$\quad$$\quad$$\quad$$\quad$$\quad$$sum_{nq}=$Cal-Seq-Dif$(n,q,ds)$;
	
	08$\quad$$\quad$$\quad$end for
	
	09$\quad$$\quad$$\quad$$T=Sort(Sum_{n})$;
	
	10$\quad$$\quad$$\quad$$\tilde{y}_{n}=T(1)$;
	
	11$\quad$$\quad$$\quad$$X.flag=0$;
	
	12$\quad$$\quad$$\quad$for $k=1:K$
	
	13$\quad$$\quad$$\quad$$\quad$$\quad$for $g=T(k)-\frac{num}{2}:T(k)+\frac{num}{2}$;
	
	14$\quad$$\quad$$\quad$$\quad$$\quad$$\quad$$\quad$$X_{g}.flag=1$;
	
	15$\quad$$\quad$$\quad$$\quad$$\quad$end for
	
	16$\quad$$\quad$$\quad$end for
	
	17$\quad$end
	\label{algorithm4}
\end{algorithm}

\begin{figure}
	\subfloat{\includegraphics[width=1\columnwidth]{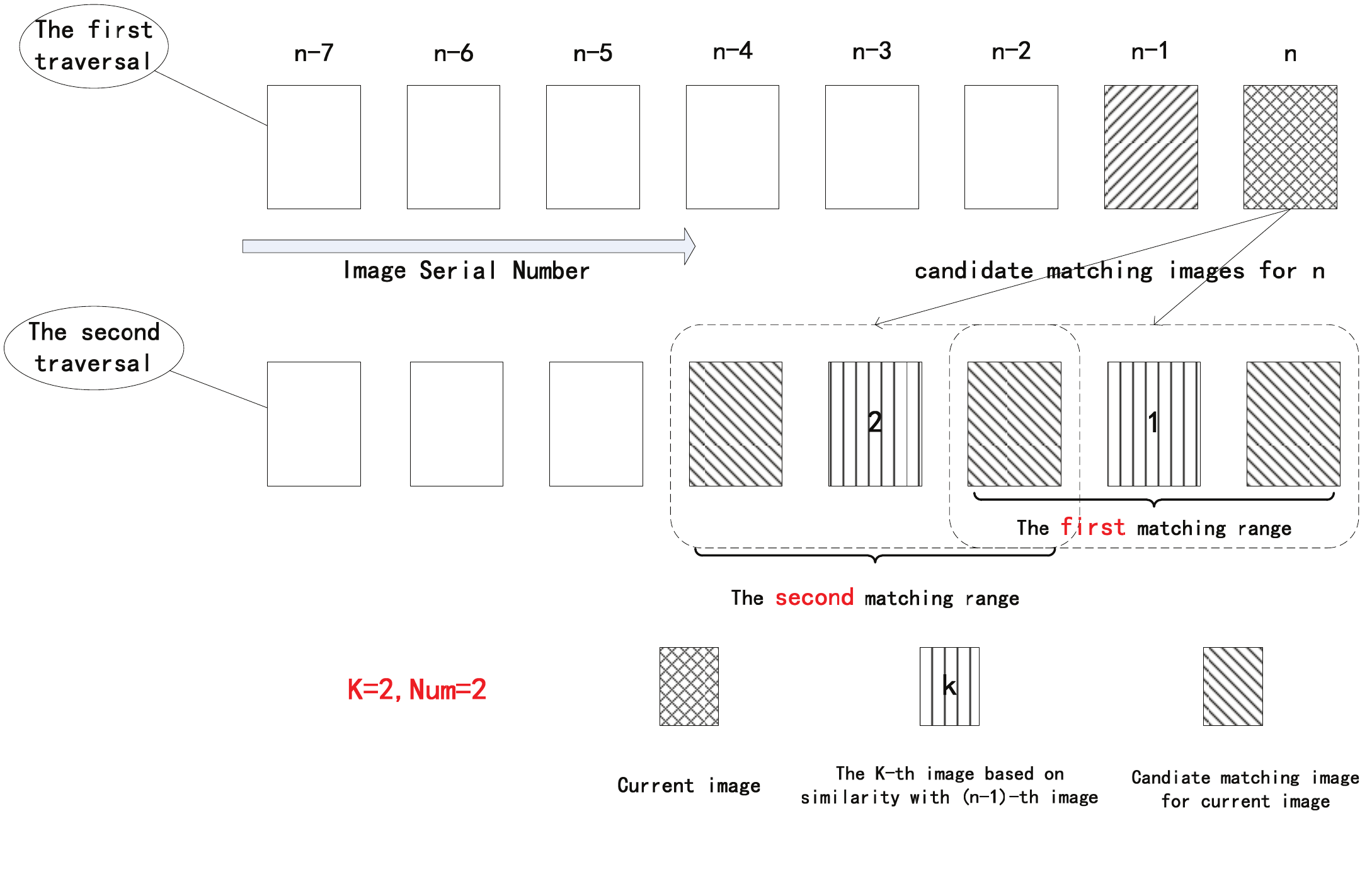}}
	
	\caption{An instance of matching ranges exists some overlaps}
	
	\label{figure8}
\end{figure}

As illustrated in steps 6 to 16 of Algorithm \ref{algorithm4}, when the image number is greater than 1 ($n$ is larger than one), the algorithm determines the matching image only from their $K$ matching ranges rather than the entire dataset. Hence, for a value of $ds$, $K$, and $Num$, regardless of the number images the dataset contains, we only need to search at most $(K(Num+1))$ times to find the matching image for an image. Besides, because of the robustness of SeqCNNSLAM (pool5), $K$ reference images are likely to be in an adjacent location, which results to some overlap in the $K$ matching ranges, such as the instance shown in Fig.\ref{figure8}. Therefore, the actual number of candidate matching images for an image is likely to be less than $(K(Num+1))$. Thus, the accelerated algorithm has good scalability and can easily deal with large-scale scenes. Additionally, in order to reduce the cumulative error, we reinitialize the algorithm every $L=450$ images by calculating the whole $(ds+1)$ columns.

\begin{figure*}
	\subfloat[]{\includegraphics[width=0.333\linewidth,height=4.3cm]{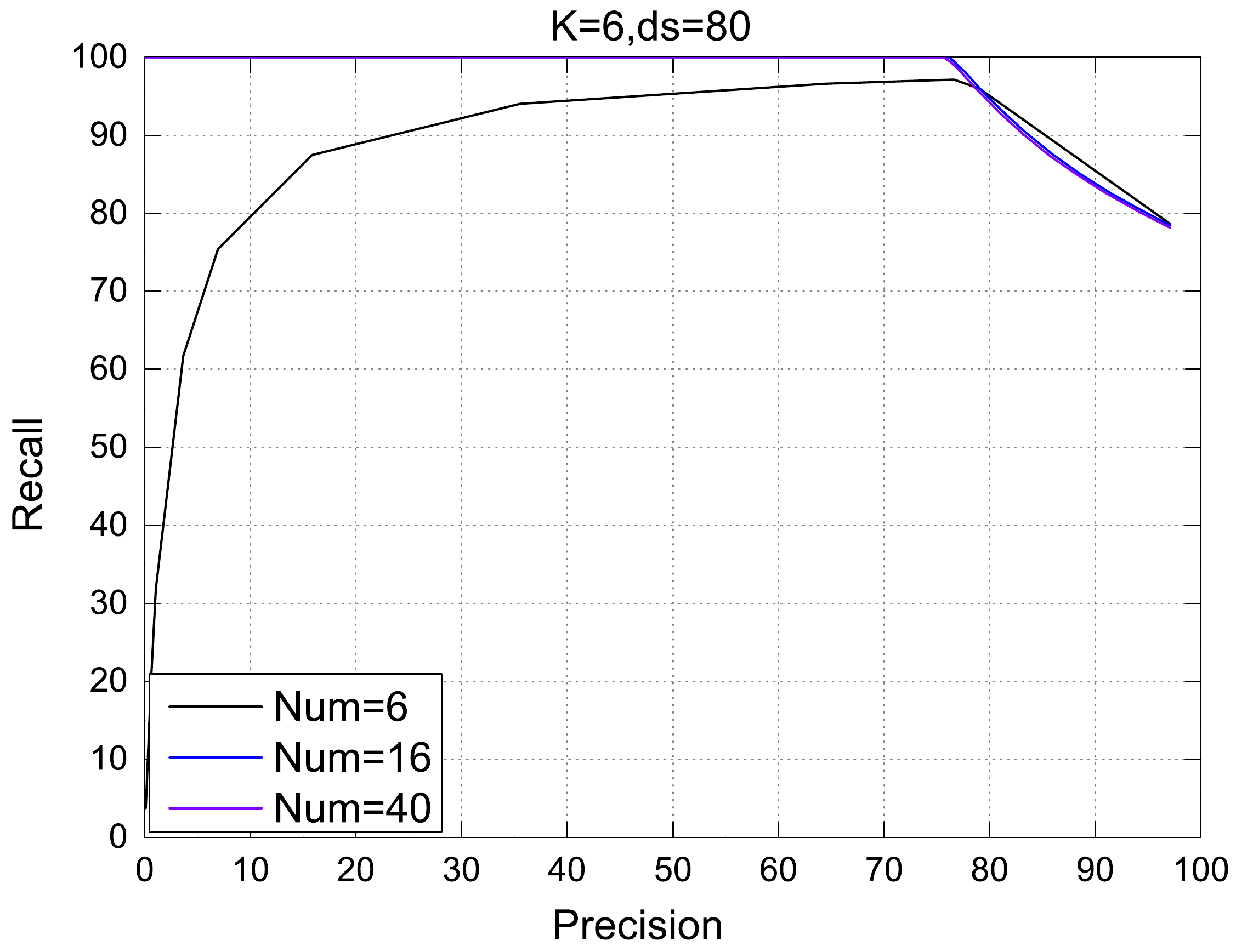}}\subfloat[]{\includegraphics[width=0.333\linewidth,height=4.3cm]{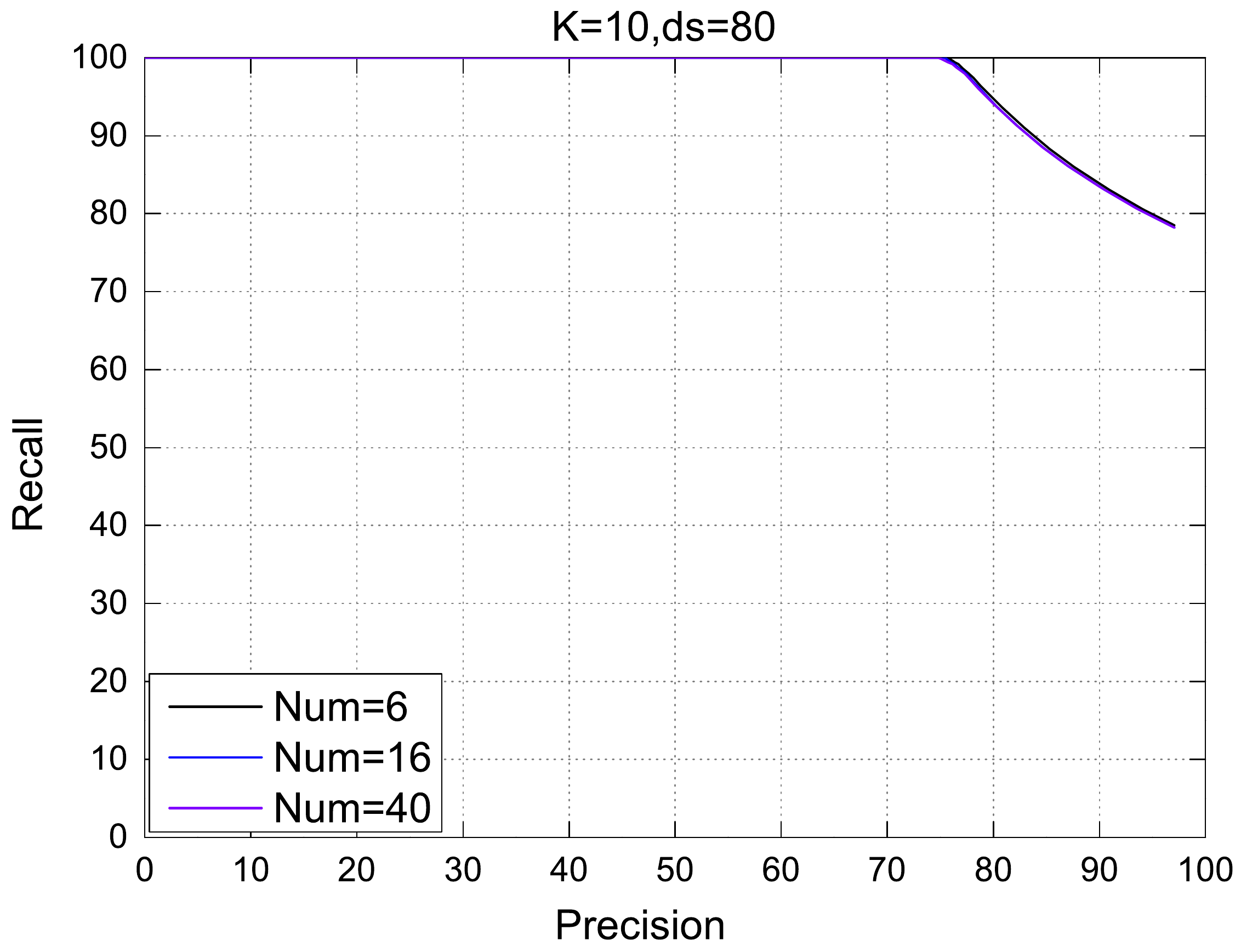}}\subfloat[]{\includegraphics[width=0.333\linewidth,height=4.3cm]{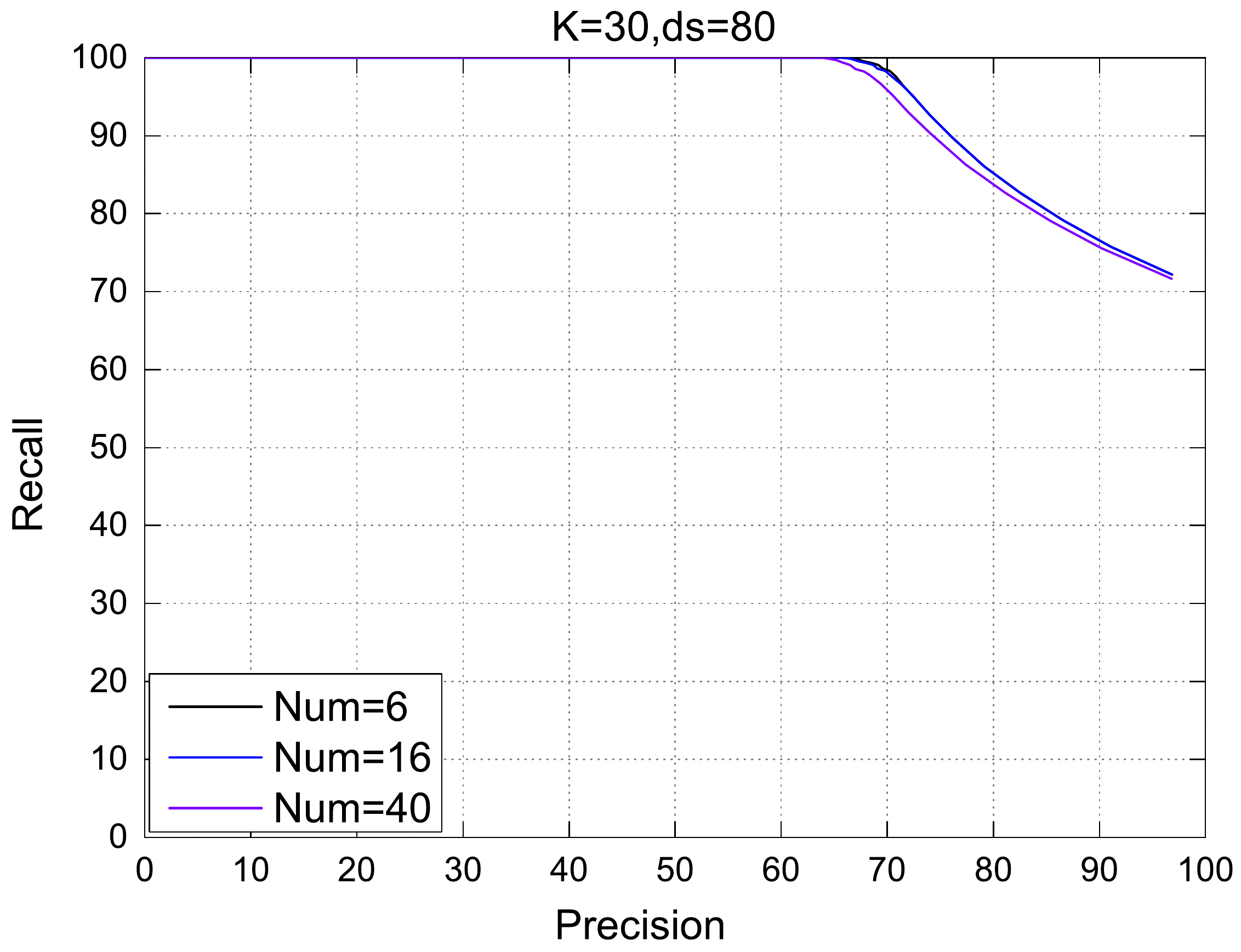}}
	
	\subfloat[]{\includegraphics[width=0.333\linewidth,height=4.3cm]{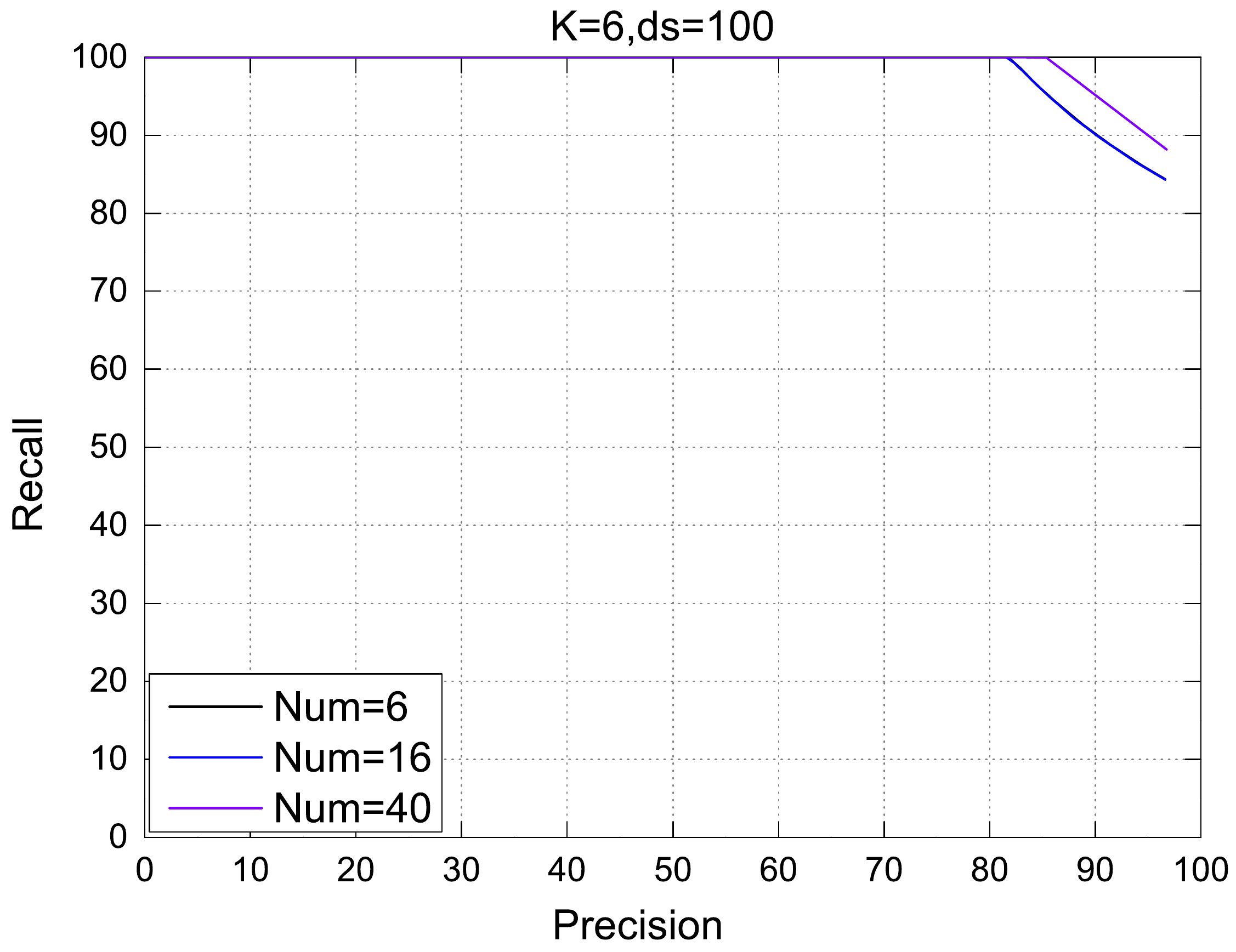}}\subfloat[]{\includegraphics[width=0.333\linewidth,height=4.3cm]{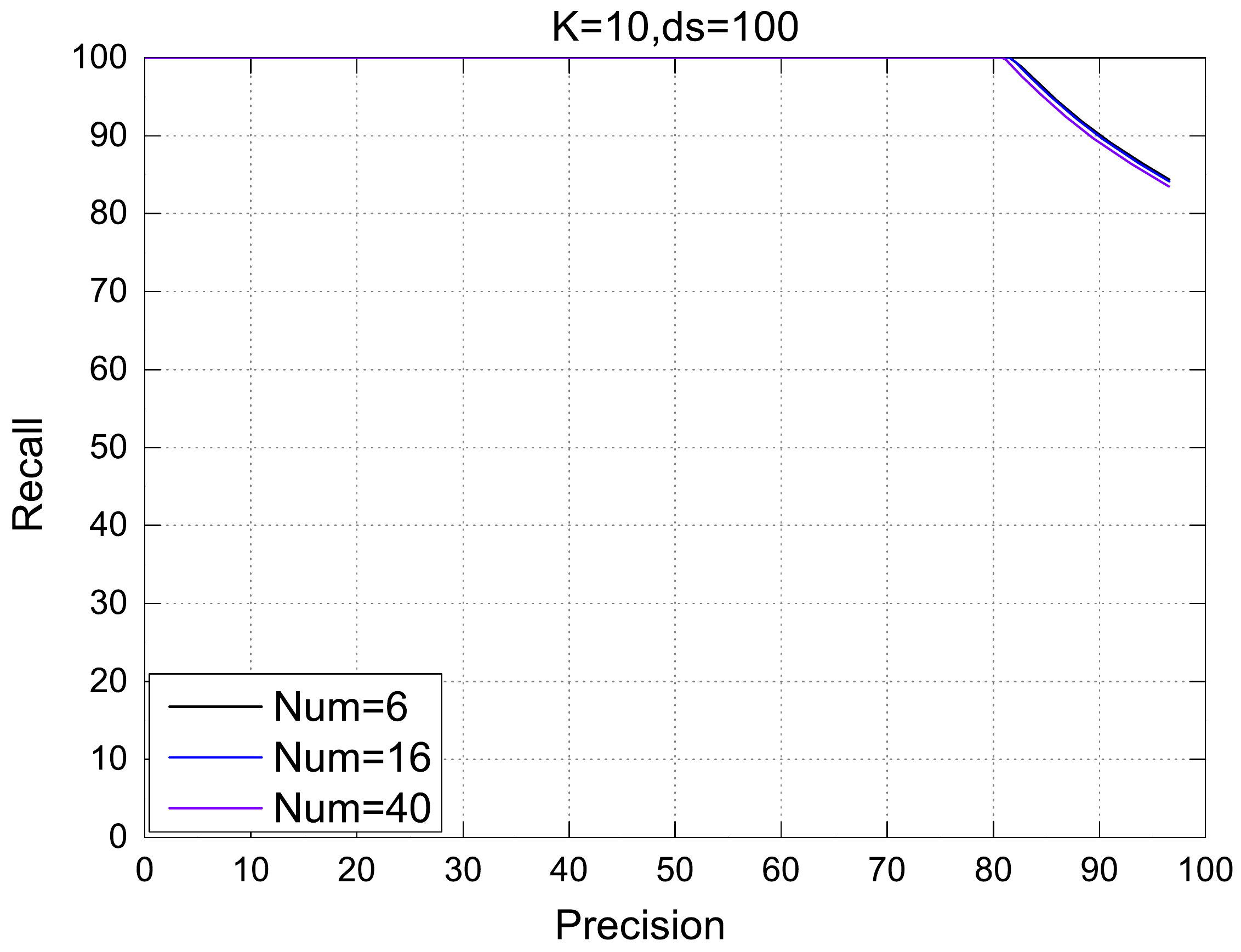}}\subfloat[]{\includegraphics[width=0.333\linewidth,height=4.3cm]{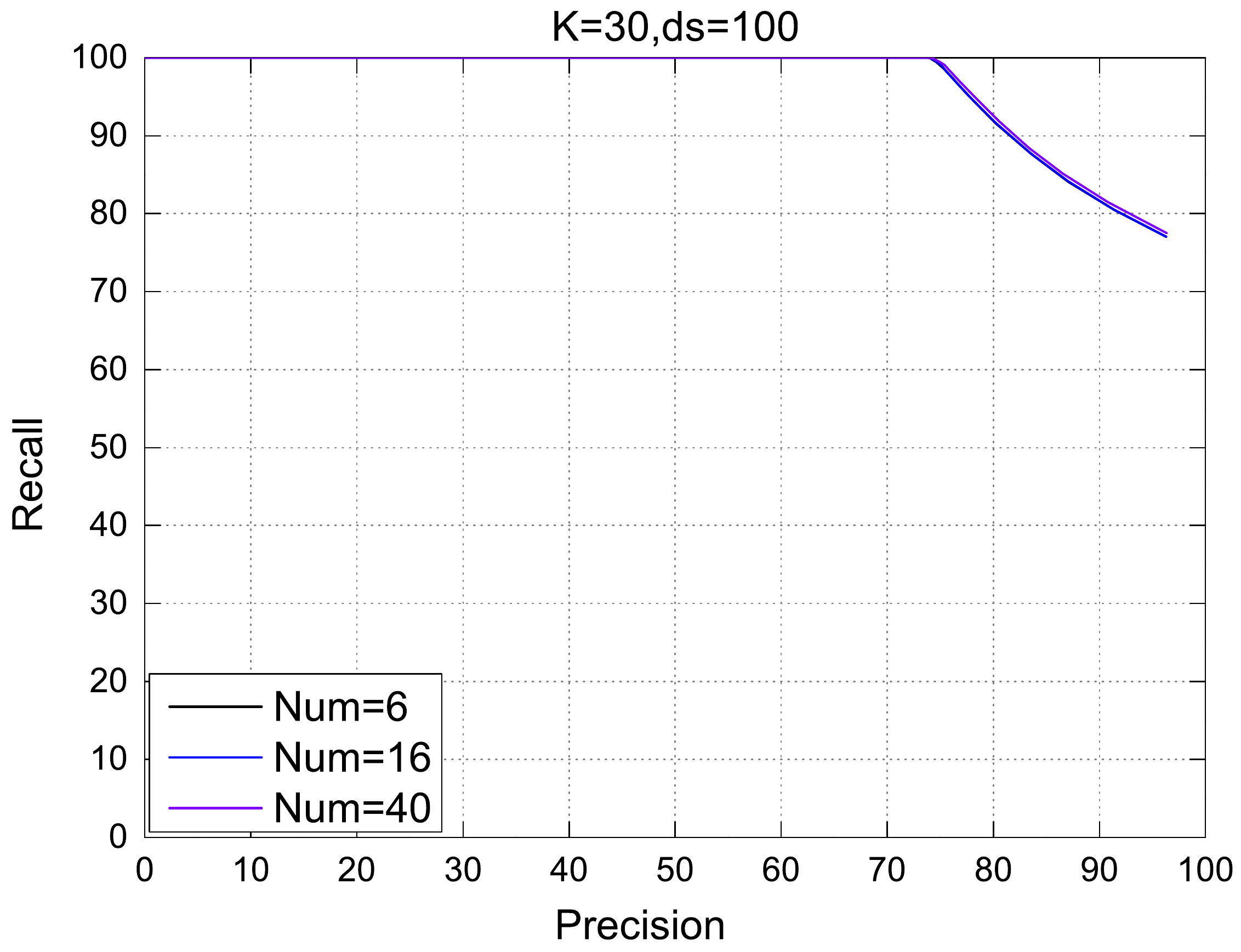}}
	
	\caption{Performance of A-SeqCNNSLAM (pool5) in the Nordland dataset with changed
		condition and changed viewpoint by 12.5\% shift and with variable
		$K$, $Num$ and $ds$. A-SeqCNNSLAM (pool5) achieves a performance that is
		comparable to that of SeqCNNSLAM (pool5) when $K$ and $Num$ are set to
		a small value, such as $K$=10 and $Num$=6.}
\label{figure9}
\end{figure*}

To verify our method, we implement it and compare its performance with that of SeqCNNSLAM(pool5) with two typical values of $ds$ (ds=80 and 100). Satisfactory performance is achieved in the Nordland dataset. Fig.\ref{figure9} shows the result of these experiments. Table 4 summarizes the runtime for the different $K$, $Num$ and $ds$. With the increment in $K$ and $Num$, the runtime of the algorithm increases, but the increment rate is gradual.

\begin{table}
	\caption{Runtime of A-SeqCNNSLAM}

	\noindent \centering{}%
	\begin{tabular}{|c|c|c|c|c|}
		\hline
		$ds$ & \backslashbox{$K$}{$Num$} & 6 & 16 & 40\tabularnewline
		\hline
		\multirow{3}{*}{80} & 6 & 98.373s  & 103.865s & 114.847s\tabularnewline
		\cline{2-5}
		& 10 & 106.392s  & 111.421s  & 122.642s\tabularnewline
		\cline{2-5}
		& 30 & 120.258s  & 129.351s & 145.465s\tabularnewline
		\hline
		\multirow{3}{*}{100} & 6 & 130.137s  & 137.691s & 147.377s\tabularnewline
		\cline{2-5}
		& 10 & 139.198s  & 141.221s & 155.016s\tabularnewline
		\cline{2-5}
		& 30 & 153.255s  & 160.601s & 183.555s\tabularnewline
		\hline
	\end{tabular}
	\label{table4}
\end{table}

The experiments consistently show that our accelerated method can achieve comparable performance even though  $K$ and $Num$ are set to a small value only. For instance, when $K=10$, $Num=6$ and $ds=100$, A-SeqCNNSLAM (pool5) and SeqCNNSLAM (pool5) exhibit consistent performances, and the best matching image among 3476 candidates for an image is identified within 40 ms on a standard desktop machine with Intel i7 processor and 8 GB memory. This condition corresponds to a speed-up factor of 4.65 using non-optimized Matlab implementation based on OpenSeqSLAM implementation\cite{Niko-2013-p}. Table \ref{table4} summarizes the required time for the main algorithmic steps. We can see that the A-SeqCNNSLAM (pool5) achieves significantly better real-time performance in large scale maps.

\subsection{O-SeqCNNSLAM: Online Learning Algorithm to Choose $K$ for A-SeqCNNSLAM}
Although we provide an efficient method to accelerate the LCD algorithm by reducing the number of candidate matching images for an image with parameters $K$ and $Num$, these two parameters are highly dataset-specific and depend largely on the trajectory of the robot. Thus, they are difficult to apply in unseen scene data or different robot configurations. A-SeqCNNSLAM  is thus not yet practical. The same dilemma of parameter selection also exists in Change Removal\cite{Lowry-2015-p} and SeqSLAM\cite{Milford-2012-p1643}\cite{Niko-2013-p}.

This section provides the O-SeqCNNSLAM method for online parameter selection that allows the A-SeqCNNSLAM algorithm to tune its parameters by observing the unknown environment directly.

Fig.\ref{figure9} shows that when $Num$ is equal or greater than 16, the performance of accelerated SeqCNNSLAM is almost marginally affect by $Num$. Hence, we can set $Num$ to a large value and provide a method to adjust parameter $K$. The historical information of an image's matching location can be exploited to solve this problem.

In A-SeqCNNSLAM, we provide $K$ matching ranges for a image, but we find that the serial number of the matching range where the best matching image is located is often less than $K$. For instance, an image A's matching image is located in its $k$-th matching range, and $k$ is less than $K$. With this clue, we provide an approach to adjust the value of $K$ online based on the location of the last few images' matching images. That is, for each image, we record the serial number of the matching range where its matching image located in, and the serial number is regard as its \emph{image matching label} (IML), corresponding to step 5 of Algorithm \ref{algorithm5}. Then, for the current image, the value of $K$ is set as the maximum value of its last few images' IML.

\begin{algorithm}
	\caption{Online Adjustment of K}

	\textbf{Require:} $S=\left\{ \left(x_{i},y_{i}\right)\right\} _{i=1}^{N}$:
	dataset for LCD containing $N$ images, $x_{i}$: input images, $y_{i}$:
	the ground truth of matching images' serial number for $x_{i}$; $\left\{ X_{i}\right\} _{i=1}^{N}$:
	CNN descriptors of $x_{i}$; $ds$: sequence length; $K$: the number
	of matching ranges for an image; $k$: serial number of a matching
	range; $Num$: the lengths of each matching sequence.
	
	\textbf{Ensure:} $K$.
	
	01$\quad$Initial: $t=0$;
	
	02$\quad$if $ChangeDegree(X_{n})\in[0.9,1.1]$
	
	03$\quad$$\quad$$\quad$for $k=1:K$
	
	04$\quad$$\quad$$\quad$$\quad$$\quad$if $\tilde{y}_{n-1}\in\left[T(k)-\frac{num}{2},T(k)+\frac{num}{2}\right]$
	
	05$\quad$$\quad$$\quad$$\quad$$\quad$$\quad$$\quad$$IML_{t}=k$;
	
	06$\quad$$\quad$$\quad$$\quad$$\quad$$\quad$$\quad$break;
	
	07$\quad$$\quad$$\quad$$\quad$$\quad$end if
	
	08$\quad$$\quad$$\quad$end for
	
	09$\quad$$\quad$$\quad$$t=t+1$;
	
	10$\quad$$\quad$$\quad$if $t==10$
	
	11$\quad$$\quad$$\quad$$\quad$$\quad$$K=max(IML)$;
	
	12$\quad$$\quad$$\quad$$\quad$$\quad$$t=0$;
	
	13$\quad$$\quad$$\quad$end if
	
	14$\quad$else
	
	15$\quad$$\quad$$\quad$$K=initial_{K}$;
	
	16$\quad$$\quad$$\quad$$g=0$;
	
	17$\quad$end if
	\label{algorithm5}
\end{algorithm}

\begin{table}
	\caption{Runtime of Online SeqCNNSLAM}

	\noindent \begin{centering}
		\begin{tabular}{|c|c|c|}
			\hline
			$ds$ & 80 & 100\tabularnewline
			\hline
			time & 121.494s & 151.932s\tabularnewline
			\hline
		\end{tabular}
		\label{table5}
		\par\end{centering}
	
\end{table}

However, when the scene changes drastically, the historical information of an image's matching location can not be used as the basis to set the $K$ value for the next batch of images. Thus, we defined a metric called \emph{Change Degree}  to measure the scene change which is defined in Eq.(4). The numerator of Eq.(4) is the sum of Euclidean distance between the current image and its last 10 images, and the denominator is the sum of Euclidean distance between the last image and its last 10 images.

\begin{equation}
ChangeDegree\left(X_{i}\right)=\frac{\sum_{i=n-10}^{n-1}Dis\left(X_{n}-X_{i}\right)}{\sum_{i=n-10}^{n-1}Dis\left(X_{n-1}-X_{i-1}\right).}
\end{equation}

Given that most steps of O-SeqCNNSLAM are identical to those of A-SeqCNNSLAM (except for the method of online adjustment of $K$), only the concrete steps about how to adjust $K$ online in Algorithm \ref{algorithm5} are provided to avoid duplication. In the first step, the change degree of the current images is calculated. If the change degree is larger than 0.9 and smaller than 1.1, the $K$ is set to maximum value of its last $t=10$ images' IML in step 11. However, if the change degree is larger than 1.1 or smaller than 0.9, $K$ is reinitialized with the initial value, as in step 15.

\begin{figure}
	\subfloat{\includegraphics[width=1\columnwidth]{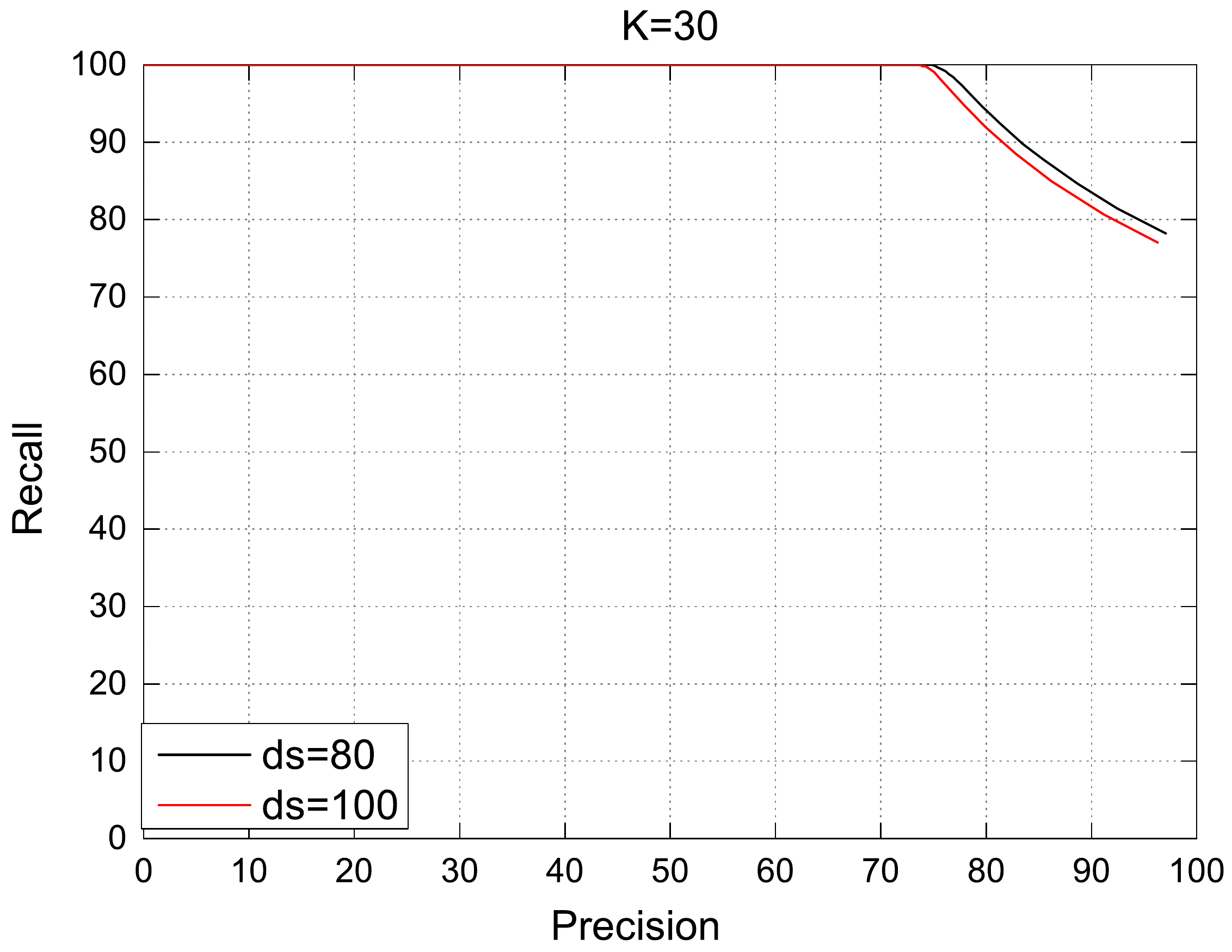}}
	
	\caption{Performance of O-SeqCNNSLAM(pool5) in Nordland dataset with changed
		conditions and changed viewpoint by 12.5\% shift. O-SeqCNNSLAM(pool5)
		achieves a performance that is comparable to that of A-SeqCNNSLAM(pool5).}
	\label{figure10}
	
\end{figure}

In order to compare with A-SeqCNNSLAM,  we also test the utility of O-SeqCNNSLAM to on the Nordland dataset with 12.5\% shift. Meanwhile, the $ds$ is set to 80 as well as 100. Fig.\ref{figure10} and Table \ref{table5} show the resulting precision-recall curves and runtime for O-SeqCNNSLAM  when $Num$ is set to 16 and the initial  $K$ is set to a large value of 30. Compared with the A-SeqCNNSLAM method, O-SeqCNNSLAM also presents a robust performance, and the runtime of the two are close. So the O-SeqCNNSLAM is an effective way to be employed for actual scene.

\section{Conclusions and Future Work}
Thorough research was conducted on the utility of SeqCNNSLAM, which is a combination of CNN features (especially pool5) and sequence matching method, for the task of LCD.  We demonstrate that directly using the pool5 descriptor can result in robust performance against combined viewpoint and condition change with the aid of SeqSLAM. A-SeqCNNSLAM was also presented to make large-scale SeqCNNSLAM possible by reducing the matching range. In O-SeqCNNSLAM, online adjustment of A-SeqCNNSLAM's parameter makes it applicable to unseen places.

In our subsequent work, we plan to apply the insights gained from this study and provide a complete method to adjust all the parameters of A-SeqCNNSLAM  simultaneously based on its operating status and provide a new distance metric to replace the Euclidean distance to avoid \emph{curse of dimensionality}\cite{Beyer-1999-p217}. Additionally, we will explore how to train a CNN specifically for LCD under combined viewpoint and condition change to improve LCD performance.

\label{}





\bibliographystyle{model3-num-names}
\bibliography{BDD_index}







\end{document}